\documentclass[conf]{IEEEtran}
\usepackage{algorithm,algpseudocode}
\usepackage{amsmath}
\usepackage{amssymb}
\usepackage{mathtools}
\usepackage[dvipsnames]{xcolor}

\usepackage{pifont}
\newcommand{\cmark}{\ding{51}}%
\newcommand{\xmark}{\ding{55}}%
\usepackage{cite}
\def\BibTeX{{\rm B\kern-.05em{\sc i\kern-.025em b}\kern-.08em
    T\kern-.1667em\lower.7ex\hbox{E}\kern-.125emX}}
\usepackage[bookmarksnumbered=true]{hyperref} 
\hypersetup{
     colorlinks = true,
     linkcolor = blue,
     anchorcolor = blue,
     citecolor = blue,
     filecolor = blue,
     urlcolor = blue
     }

\usepackage{graphicx}
\usepackage{svg}

\usepackage{wrapfig}
\usepackage{caption}
\usepackage{subcaption}
\usepackage{booktabs}

\usepackage{tikz}

\usepackage{ragged2e}
\usepackage{siunitx}
\usepackage{multicol}
\usepackage{multirow}

\usepackage{xcolor}
\begin{document}

\title{SpikeSim: An end-to-end Compute-in-Memory Hardware Evaluation Tool for Benchmarking Spiking Neural Networks}

\author{ Abhishek Moitra$^*$, \textit{Student Member, IEEE}, Abhiroop Bhattacharjee$^*$, \textit{Student Member, IEEE}, Runcong Kuang, Gokul Krishnan, \textit{Member, IEEE}, Yu Cao, \textit{Fellow, IEEE}, and Priyadarshini Panda, \textit{Member, IEEE}
\thanks{$^*$ These authors have contributed equally to this work.}
\thanks{Abhishek Moitra, Abhiroop Bhattacharjee, and Priyadarshini Panda are with
the Department of Electrical Engineering, Yale University, New Haven, CT,
USA. }
\thanks{Runcong Kuang, Gokul Krishnan, and Yu Cao are with the School of Electrical, Computer, and Energy
Engineering, Arizona State University, Tempe 85287, AZ. }
}

\maketitle

\begin{abstract}
Spiking Neural Networks (SNNs) are an active research domain towards energy efficient machine intelligence. Compared to conventional artificial neural networks (ANNs), SNNs use temporal spike data and bio-plausible neuronal activation functions such as Leaky-Integrate Fire/Integrate Fire (LIF/IF) for data processing. However, SNNs incur significant dot-product operations causing high memory and computation overhead in standard von-Neumann computing platforms. To this end, In-Memory Computing (IMC) architectures have been proposed to alleviate the ``memory-wall bottleneck" prevalent in von-Neumann architectures. Although recent works have proposed IMC-based SNN hardware accelerators, the following key implementation aspects have been overlooked 1) the adverse effects of crossbar non-ideality on SNN performance due to repeated analog dot-product operations over multiple time-steps 2) hardware overheads of essential SNN-specific components such as the LIF/IF and data communication modules. To this end, we propose SpikeSim, a tool that can perform realistic performance, energy, latency and area evaluation of IMC-mapped SNNs. SpikeSim consists of a practical monolithic IMC architecture called SpikeFlow for mapping SNNs. Additionally, the non-ideality computation engine (NICE) and energy-latency-area (ELA) engine performs hardware-realistic evaluation of SpikeFlow-mapped SNNs. Based on 65nm CMOS implementation and experiments on CIFAR10, CIFAR100 and TinyImagenet datasets, we find that the LIF/IF neuronal module has significant area contribution ($>11\%$ of the total hardware area). To this end, we propose SNN topological modifications that leads to $1.24\times$ and $10\times$ reduction in the neuronal module's area and the overall energy-delay-product value, respectively. Furthermore, in this work, we perform a holistic comparison between IMC implemented ANN and SNNs and conclude that lower number of time-steps are the key to achieve higher throughput and energy-efficiency for SNNs compared to 4-bit ANNs. The code repository for the SpikeSim tool will be made available in \href{https://github.com/Intelligent-Computing-Lab-Yale/Quanitzation-aware-SNN-training-and-hardware-evaluation-for-IMC-Architectures}{this Github link.}
\end{abstract}


\IEEEpeerreviewmaketitle

\section{Introduction}

In the last decade, Spiking Neural Networks (SNNs) have gained significant attention in the context of energy-efficient machine intelligence \cite{roy2019towards}. SNNs encode input data information with discrete binary spikes over multiple time-steps making them highly suitable for asynchronous event-driven input processing applications \cite{stromatias2017event, kim2021optimizing}. Recent works have proposed full-scale general-purpose von-Neumann architectures leveraging the temporal processing property of SNNs \cite{davies2018loihi, akopyan2015truenorth}. Other works such as \cite{narayanan2020spinalflow, lee2022parallel} have proposed novel dataflow to minimize the hardware overhead in von-Neumann implementation of SNNs. However, SNNs like conventional Artificial Neural Networks (ANNs) entail significant dot-product operations leading to high memory and energy overhead when implemented on traditional von-Neumann architectures (due to the ``memory wall bottleneck") \cite{kim2012functional, ankit2017resparc}. To this end, analog In-Memory Computing (IMC) architectures \cite{ni2017energy, chakraborty2020pathways, ankit2019puma} have been proposed to perform analog dot-product or Multiply-and-Accumulate (MAC) operations to achieve high memory bandwidth and compute parallelism, thereby overcoming the ``memory wall bottleneck".
\begin{table}[t]
    \centering
    \caption{Table showing qualitative comparison of SpikeSim with related works. I- Inference, T- Training, VN- von-Neumann, IMC- In-memory Computing, ELA- Energy, Latency \& Area, M- Monolithic and C- Chiplet Architecture.}
    \resizebox{\linewidth}{!}{
    \begin{tabular}{|l|l|c|c|c|c|}
    \hline
     & \multirow{2}{*}{\textbf{Work}} & \multirow{2}{*}{\textbf{Platform}} & \multirow{2}{*}{\textbf{I / T}} & \multirow{2}{*}{\begin{tabular}{c}
           \textbf{Non-} \\ \textbf{Ideality}
     \end{tabular}} & \multirow{2}{*}{\begin{tabular}{c}
           \textbf{ELA} \\ \textbf{Evaluation}
     \end{tabular}} \\ 
     &  &  & & &  \\ \hline

     \multirow{5}{*}{\rotatebox[origin=c]{90}{\textbf{ANN}}} & Eyeriss \cite{eyeriss} & VN-M & I & \xmark & \cmark \\ \cline{2-6}
     & Neurosim \cite{chen2018neurosim} & IMC-M & I & \xmark & \cmark \\ \cline{2-6}
     & CrossSim \cite{plimpton2016crosssim} & IMC-M &  I & \xmark & \cmark \\ \cline{2-6}
     & RxNN \cite{jain2020rxnn} & IMC-M & I & \cmark & \xmark \\ \cline{2-6}
     &SIAM \cite{krishnan2021siam} & IMC-C & I & \cmark & \cmark \\ \hline
     \multirow{5}{*}{\rotatebox[origin=c]{90}{\textbf{SNN}}} & Loihi \cite{davies2018loihi}, TrueNorth \cite{akopyan2015truenorth} & VN-M & I & \xmark & \xmark \\ \cline{2-6}
     & SpinalFlow \cite{narayanan2020spinalflow}, PTB \cite{lee2022parallel}
     & {VN-M} & {I} & {\xmark} & {\xmark} \\ \cline{2-6}
     & H2Learn \cite{liang2021h2learn}, SATA \cite{yin2022sata} & VN-M & T & \xmark & \cmark \\ \cline{2-6}
     & RESPARC \cite{ankit2017resparc} & IMC-M & I & \xmark & \xmark \\ \cline{2-6}
     & \textbf{SpikeSim (ours)} & \textbf{IMC-M} & \textbf{I} & \cmark & \cmark \\ \hline
    \end{tabular}}
    
    \label{tab:my_label}
\end{table}

Being an emerging and heavily researched computing paradigm, IMC architectures require hardware evaluation platforms for fast and accurate algorithm benchmarking. To this effect, many state-of-the-art hardware evaluation frameworks \cite{chen2018neurosim, jain2020rxnn, krishnan2021siam, plimpton2016crosssim} have been proposed for realistic evaluation of IMC-mapped ANNs. However, they are unsuitable for hardware-realistic SNN evaluations as they lack key architectural modifications required for temporal spike processing and non-linear activation functions, such as Leaky Integrate Fire or Integrate Fire (LIF/IF). In the context of hardware evaluation platforms for SNNs, works such as \cite{yin2022sata, liang2021h2learn} have been proposed for benchmarking SNN training on digital CMOS platforms. Additionally, works such as \cite{ankit2017resparc} propose IMC architectures for SNN inference. However, they lack several practical architectural considerations such as non-idealities incurred during analog MAC computations \cite{chakraborty2020geniex, krishnan2021robust, krishnan2022exploring}, data communication overhead among others rendering them unsuitable for a holistic hardware evaluation for IMC mapped SNNs. All of these have been qualitatively illustrated and compared in Table \ref{tab:my_label}. Therefore, in current literature, there is an evident gap between SNN algorithm design and a holistic evaluation platform for hardware-realistic benchmarking of these algorithms.



To this end, we propose \textit{SpikeSim}, an end-to-end hardware evaluation tool for benchmarking SNN inference algorithms. SpikeSim consists of a monolithic IMC-based tiled hardware architecture called \textit{SpikeFlow} that maps a given SNN on non-ideal analog crossbars. In SpikeFlow, we incorporate SNN-specific non-linear activation functions such as LIF/IF neuron and leverage the binary spike input data to propose a lightweight  module (the \textit{DIFF} module) for facilitating signed MAC operations without the need for traditional dual-crossbar approach \cite{chen2018neurosim,truong2014new}. For hardware-realistic SNN inference performance benchmarking, we develop a Non-Ideality Computation Engine (NICE). NICE incorporates a non-ideality-aware weight encoding to improve the robustness of SNNs when mapped on analog crossbars \cite{bhattacharjee2022examining}. NICE incorporates circuit analysis methods to realize non-ideal MAC operations and provide hardware realistic SNN inference performance. Furthermore, we design an \textit{Energy-Latency-Area (ELA)} engine to benchmark hardware realistic energy, latency and area of the SpikeFlow-mapped SNN. 


The key contributions of our work can be summarized as follows:
\begin{enumerate}
    \item We propose SpikeSim which is an end-to-end hardware benchmarking tool for SNN inference. SpikeSim consists of SpikeFlow- a tiled memristive crossbar architecture. SpikeFlow incorporates Leaky-Integrate-Fire/ Integrate-Fire (LIF/IF) functionality, and a novel fully-digital DIFF module that eliminates dual-crossbar approach for signed MAC computations \cite{truong2014new, chen2018neurosim}. Additionally, it contains NICE and ELA engines for crossbar-realistic hardware evaluations. 
    
    \item We develop NICE to perform fast and realistic modelling of resistive and device conductance variation non-idealities for crossbar-aware performance evaluations of SNNs. NICE incorporates a non-ideality aware weight encoding scheme that improves the inference accuracy of pretrained SNNs implemented on analog crossbars.
    
     \item We perform extensive hardware evaluations on benchmark datasets- CIFAR10, CIFAR100 \cite{cifar}, TinyImagenet \cite{le2015tiny} and unravel that the neuronal module consumes a significant portion of the total chip area ($11-30\%$) owing to the requirement to store a large number of membrane potentials in between time-steps. 
     
     \item Through extensive experiments we show that simple SNN topological modifications, such as reducing the number of output channels in the first convolutional layer, can ameliorate the area overhead of the neuron module by $1.24\times$ and improve the Energy-Delay Product (EDP) by $ 10\times$. Furthermore, we show that the non-ideality aware weight encoding improves the crossbar-mapped SNN accuracy by more than $70\%$ (for CIFAR10 dataset) compared to vanilla weight encoding onto the SpikeFlow architecture. 
     
     \item Finally, we compare the performance as well as area and energy distributions of crossbar mapped VGG9 ANN and SNNs trained on CIFAR10 dataset. We find that SNNs exhibit $\sim1000 \times$ higher neuronal module area compared to ANNs and can achieve iso-performance and higher energy-efficiency and throughput benefits at small value of time-steps ($T$=3,4,5) compared to 4-bit ANNs. 
    

\end{enumerate}

To the best of our knowledge, SpikeSim is the first hardware-realistic evaluation platform for SNNs mapped on IMC architecture. Through SpikeSim, we bring out some of the key parameters in SNN algorithm and IMC architecture design that can potentially lead to IMC-aware SNN research directions in the future.

\section{Related Works}
\subsection{Hardware Evaluation Platforms for ANN Inference}
Eyeriss \cite{eyeriss} has proposed a reconfigurable digital systolic-array architecture for energy-efficient ANN accelerators. The authors show that data transfer from DRAM memory to the computation unit contributes significantly to the energy consumption in von-Neumann ANN accelerators and hence propose a row-stationary dataflow to mitigate the memory overhead. More recent works such as ISAAC \cite{shafiee2016isaac}, used in-memory computing architectures such as analog crossbars to perform fast and energy efficient computation of ANNs. They performed extensive hardware evaluation with different crossbar sizes, analog-to-digital converter (ADC) precision among others. PUMA \cite{ankit2019puma} proposes a memristive crossbar-based ANN accelerator that uses graph partitioning and custom instruction set architecture to schedule MAC operations in a multi-crossbar architecture. The work by Chen et al. \cite{chen2018neurosim} Neurosim, proposes an end-to-end hardware evaluation platform for evaluating monolithic analog crossbar-based ANN accelerators. Recent work SIAM by Krishnan et al. \cite{krishnan2021siam} proposed an end-to-end hardware evaluation platform for chiplet-based analog crossbar-based ANN accelerators.
While the above works provide state-of-the-art evaluation platforms for ANN accelerators, they are insufficient for accurate SNN evaluation as they lack critical architectural modifications required for temporal spike data processing and LIF/IF activation functionalities. 

\subsection{Hardware Evaluation Platforms for SNN Inference}
In a recent work SpinalFlow \cite{narayanan2020spinalflow}, Narayanan et al. showed that naive hardware implementation of SNNs on Spiking Eyeriss-like architecture lowers the energy-efficiency claimed by SNNs. To this end, the work proposed architectural changes and used a tick-batched dataflow to achieve higher energy efficiency and lower hardware overheads. Another work RESPARC \cite{ankit2017resparc} proposed analog crossbar-based hardware accelerators for energy efficient implementation of SNNs. The energy efficiency of their implementation is achieved due to the event-driven communication and computation of spikes. However, the work overlooks the underlying hardware overheads for event-driven communication and the effect of analog crossbar non-idealities on SNN performance.


Given the current literature gap in IMC-based hardware evaluation platforms for SNNs, we propose SpikeSim, an end-to-end platform for hardware realistic benchmarking of SNNs implemented on IMC architectures. SpikeSim contains SpikeFlow crossbar architecture that incorporates SNN-specific spike data processing and LIF/IF Neuron functionality. SpikeSim also incorporates the NICE and ELA engine for hardware-realistic performance, energy, latency and area evaluation of IMC-mapped SNNs.


\section{Background}

\subsection{Spiking Neural Networks}

SNNs \cite{roy2019towards,diehl2015unsupervised} have gained attention due to their potential energy-efficiency compared to standard ANNs. 
The main feature of SNNs is the type of neural activation function for temporal signal processing, which is different from a ReLU activation for ANNs.
A Leak-Integrate-and-Fire (LIF) neuron   is commonly used as an activation function for SNNs.
The LIF neuron $i$ has a membrane potential $u_{i}^{t}$ which accumulates the weighted summation of asynchronous spike inputs $S_j^{t}$, which can be formulated as follows:
\begin{equation}
    U_i^t = \lambda  U_i^{t-1} + \sum_j w_{ij}S^t_j.
    \label{eq:LIF}
\end{equation}
Here, $t$ stands for time-step, and $w_{ij}$ is for weight connections between neuron $i$ and neuron $j$. Also, $\lambda$ is a leak factor.
The LIF neuron $i$ accumulates membrane potential and generates a spike output $o_i^{t}$ whenever membrane potential exceeds the threshold $\theta$:
\begin{equation}
    o^{t}_i =
\begin{cases}
 1,          & \text{if $u_i^{t} >\theta$},  \\
    0
    & \text{otherwise.} 
\end{cases}
\label{eq:firing}
\end{equation}
The membrane potential is reset to zero after firing.
This integrate-and-fire behavior of an LIF neuron generates a non-differentiable function, which is difficult to be used with standard backpropagation.

To address the non-differentiability, various training algorithms for SNNs have been studied in the past decade.
ANN-SNN conversion methods \cite{sengupta2019going,diehl2015fast,han2020deep,li2021free,rueckauer2017conversion} convert pretrained ANNs to SNNs using weight (or threshold) scaling   in order to approximate ReLU activation with LIF/IF activation.
They can leverage well-established ANN training methods, resulting in high accuracy on complex datasets.
On the other hand, surrogate gradient learning addresses the non-differentiability problem of an LIF/IF neuron by approximating the backward gradient function \cite{wu2018spatio}.
Surrogate gradient learning can directly learn from the spikes, in a smaller number of time-steps. 

Based on the surrogate learning, several input data encoding schemes have been compared. A recent work \cite{kim2022rate} compares two state-of-the-art input data encoding techniques- {Direct Encoding} and {Rate Encoding}. Rate encoding converts a input data to stochastically distributed temporal spikes using poisson coding technique \cite{bntt}. In contrast, direct encoding leverages features directly extracted from the inputs over multiple time-steps. It has been shown that direct encoding schemes can achieve higher performance at lower number of time-steps.


\subsection{Analog Crossbar Arrays and their Non-idealities}

\begin{wrapfigure}{l}{0.3\textwidth}
\includegraphics[width=0.3\textwidth]{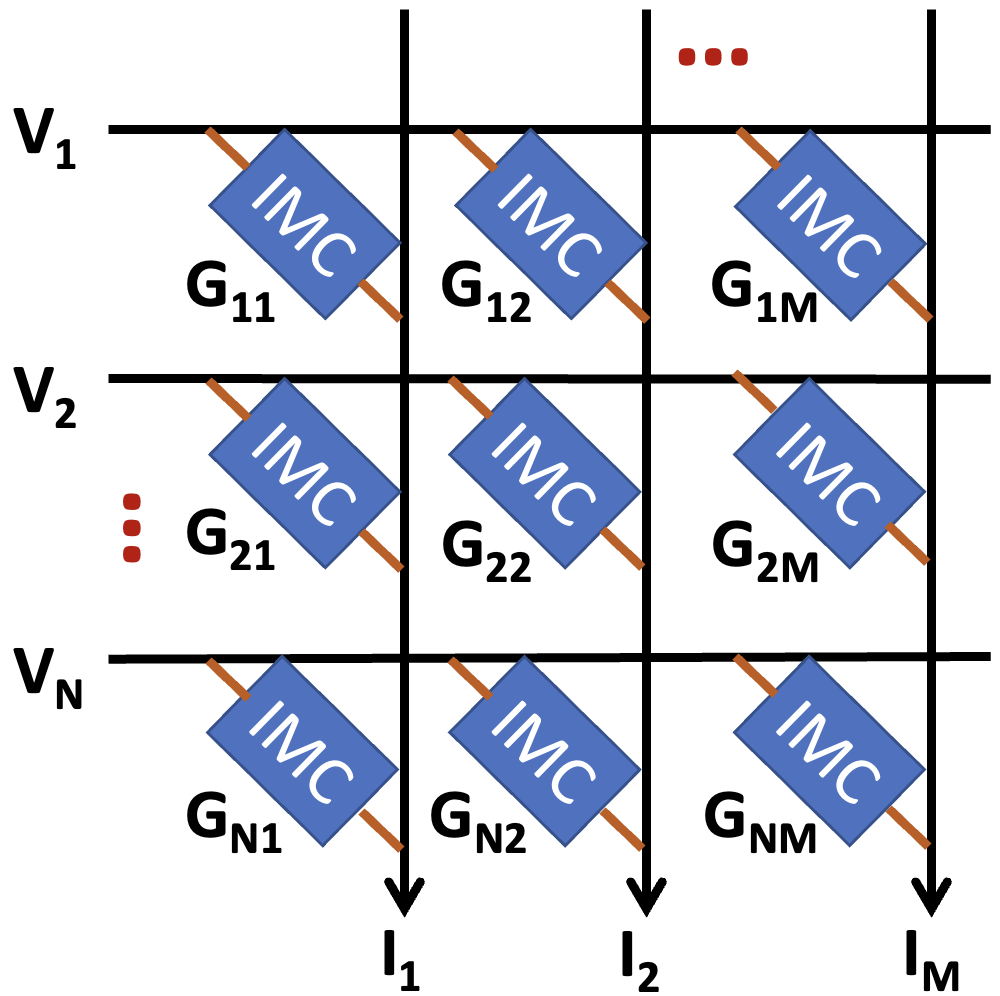}
  
\caption{An IMC crossbar array with input voltages $V_{i}$, IMC devices bearing synaptic conductances $G_{ij}$ and output currents $I_{j}$.}
\label{xbar_img}

\end{wrapfigure}

Analog crossbars consist of 2D arrays of In-Memory-Computing (IMC) devices, Digital-to-Analog Converters (DACs) and Analog-to-Digital Converters (ADCs) and write circuits for programming the IMC devices.
The activations of a neural network are fed in as analog voltages $V_i$ to each row of the crossbar and weights are programmed as synaptic device conductances ($G_{ij}$) at the cross-points as shown in Fig. \ref{xbar_img}. For an ideal N$\times$M crossbar during inference, the voltages interact with the device conductances and produce a current (governed by Ohm's Law). 

Consequently, by Kirchoff's current law, the net output current sensed at each column $j$ is the sum of currents through each device, \textit{i.e.} $I_{j(ideal)} = \Sigma_{i=1}^{N}{G_{ij} * V_i}$. 
We term the matrix $G_{ideal}$ as the collection of all $G_{ij}$'s for a crossbar. However, in reality, the analog nature of the computation leads to various hardware noise or non-idealities, such as, interconnect parasitic resistances and synaptic device-level variations \cite{jain2020rxnn, liu2015vortex, chakraborty2020geniex, bhattacharjee2021neat, bhattacharjee2022examining}. This results in a $G_{non-ideal}$ matrix, with each element $G_{ij}'$ incorporating the impact of the non-idealities. Consequently, the net output current sensed at each column $j$ in a non-ideal scenario becomes $I_{j(non-ideal)} = \Sigma_{i=1}^{N}{G_{ij}' * V_i}$, which deviates from its ideal value. This manifests as huge accuracy losses for neural networks mapped onto crossbars. Larger crossbars entail greater non-idealities, resulting in higher accuracy losses \cite{jain2020rxnn, chakraborty2020geniex, bhattacharjee2021efficiency, bhattacharjee2022examining}. 


\section{SpikeSim}

\begin{figure}[h]
    \centering
    \includegraphics[width=0.9\linewidth]{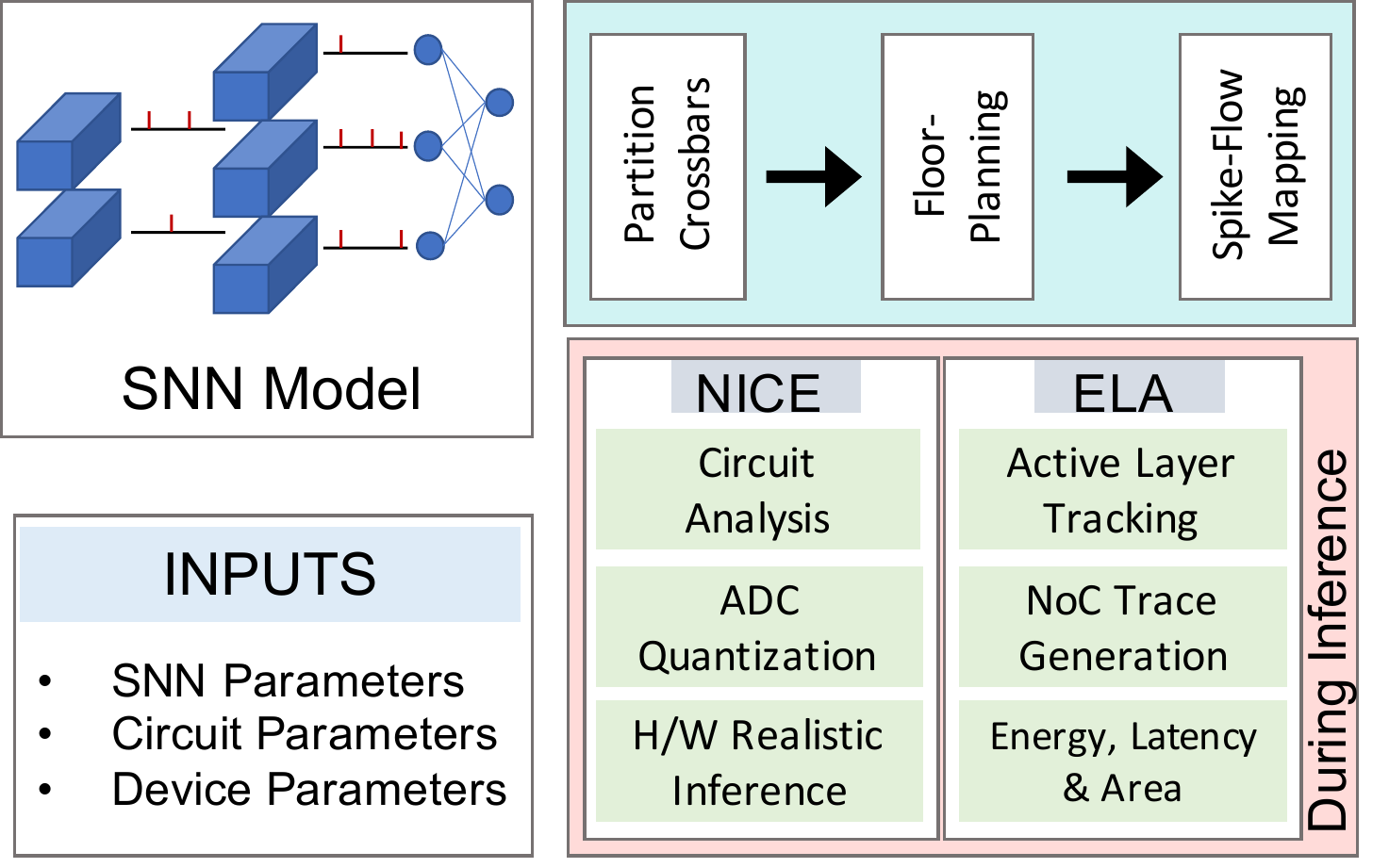}

    \caption{SpikeSim entails (a) SNN mapping on the SpikeFlow hardware archictecture (b) hardware realistic performance evaluation using NICE and (c) hardware evaluation using ELA engine.}
    \label{fig:spikesim_flow}
     
\end{figure}
\begin{table}[]
    \centering
    \caption{Table describing various SNN-level, circuit-level and device-level parameters pertaining to SpikeSim.}
    \resizebox{\linewidth}{!}{
    \begin{tabular}{|l|c|} \hline
    \multicolumn{2}{|c|}{\textbf{SNN Parameters}} \\ \hline
       Network Topology  &  SNN network structure information\\ \hline
       Sparsity & SNN layer-specific Spike Sparsity \\ \hline
       $k$ & Weight Quantization\\ \hline
       $k_{mem}$&  Membrane Potential Quantization \\ \hline
       $T$ & Number of Time-Steps \\ \hline
       Activation Type & LIF or IF \\ \hline
       \multicolumn{2}{|c|}{\textbf{Circuit Parameters}} \\ \hline
       NoC Topology & Mesh or Tree type \\ \hline
       $SU$ & DIFF module SpeedUp\\ \hline
       Scheduling Factor & Layer Scheduling Factor \\ \hline
       Clock Frequency & Frequency of operation \\ \hline 
       $X$ & IMC Crossbar Array Size\\ \hline
       $N_C$ & Crossbar count in each PE \\ \hline
       $N_{PE}$ & PE count in each Tile \\ \hline
       $MUX ~Size$ & Number of columns multiplexed \\ \hline
       $B_{GB}$, $B_{TB}$, $B_{PB}$ & Global, Tile and PE Buffer Size \\ \hline
       $B_{TIB}$, $B_{PIB}$ & Tile and PE Input Buffer Size \\ \hline
       NoC Width & NoC Channel Width \\ \hline
       $V_{DD}$ & Supply Voltage \\ \hline
       $V_{read}$ & {Read Voltage}\\ \hline
       $h$ & Precision of the Crossbar ADC \\ \hline
       
        $r$ & Column Parasitic Resistance \\ \hline
       \multicolumn{2}{|c|}{\textbf{Device Parameters}} \\ \hline
       Technology & CMOS technology \\ \hline
       IMC device & SRAM or RRAM device \\ \hline
       Bits/Cell & Precision of 1 IMC device \\ \hline
       $R_{on}$ and $R_{off}$ &  On and Off IMC device resistances\\ \hline
       $\sigma$ & Synaptic Conductance Variation \\
       \hline
       
    \end{tabular}}
    
    \label{tab:spikesim_parameters}
\end{table}
SpikeSim platform as shown in Fig. \ref{fig:spikesim_flow} requires various SNN, circuit and device parameter inputs (details provided in Table \ref{tab:spikesim_parameters}) for the hardware evaluation. It consists of three different stages:
\begin{enumerate}
    \item \textbf{SpikeFlow Mapping:} A pre-trained SNN is partitioned and mapped on a realistic analog crossbar architecture called SpikeFlow (See Section \ref{sec:spikeflow} for details).
    \item \textbf{Non-Ideality Computation Engine (NICE):} Incorporates circuit analysis and ADC quantization to evaluate hardware-realistic inference performance of SpikeFlow mapped SNNs (See Section \ref{sec:NICE}).
    \item \textbf{ELA Engine:} Computes the energy, latency and area of the SpikeFlow-mapped SNN (see Section \ref{sec:ela_engine}). 
    
\end{enumerate}

\subsection{SpikeFlow Architecture}
\label{sec:spikeflow}

\begin{figure*}[t]
    \centering
    \includegraphics[width=\linewidth]{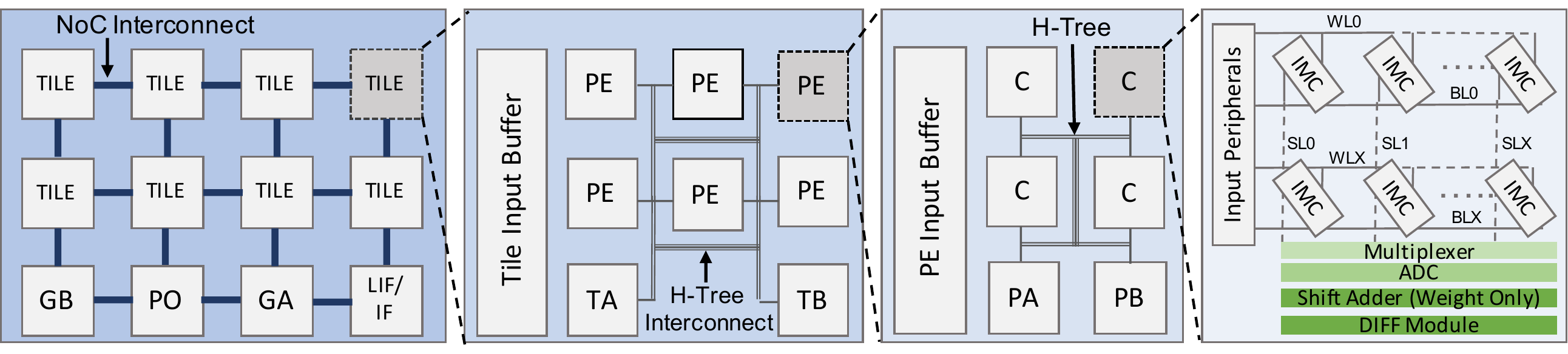}

    \caption{The hierarchical SpikeFlow architecture consisting of Tiles, Processing Elements (PEs) and analog crossbars (C). The crossbars consist of IMC device array, multiplexer, ADC, shift-adders, DIFF modules and input peripherals for realizing analog dot product operations. \textit{GB}, \textit{TB}, \textit{PB} denote global, tile and PE buffers, respectively. Similarly, \textit{GA}, \textit{TA}, \textit{PA} denote the global, tile and PE accumulators, respectively. \textit{PO}- pooling module. The LIF/IF Neuronal Module (LIF/IF) and the \textit{DIFF} Module are specific to the SpikeFlow architecture to leverage critical functions to implement SNNs. Additionally, SpikeFlow uses Network on Chip and H-Tree interconnects for inter-tile and intra-tile communications, respectively.}
    \label{fig:spikeflow}
     
\end{figure*}

\textbf{Architecture Overview: }SpikeFlow (shown in Fig. \ref{fig:spikeflow}) follows a hierarchical analog crossbar-based monolithic chip architecture \cite{chen2018neurosim}. The top hierarchy consists of Tiles and digital peripheral modules such as the global buffer (GB), pooling module (PO), global accumulator (GA) and LIF/IF neuronal module for storage, pooling, accumulation and LIF/IF neuronal activation functionality, respectively. The Tiles and peripheral modules are connected by a Network on Chip (NoC) interconnect \cite{krishnan2021siam}. Each Tile consists of a Tile Input Buffer, a fixed number of Processing Elements (PEs) and peripherals- Tile Accumulator (TA), and Tile Buffer (TB). Each PE consists of a PE Input Buffer, a fixed number of analog crossbars ($C$) and peripherals- PE Accumulator (PA) and PE Buffer (PB). Inside the Tile and PE, all modules are connected using an H-Tree 
- point to point interconnect \cite{chen2018neurosim}. The PB, TB and GB store the MAC outputs from the crossbar, PE and Tile, respectively. Similarly, PA, TA and GA accumulate the partial sum outputs from the crossbars, PE and Tile, respectively. 

Each crossbar, consists of an X$\times$X IMC device array, Input Peripherals, Multiplexers, Analog to Digital Converters (ADC), Shift-and-Add circuit and DIFF modules. Note, that the DIFF and LIF/IF Neuronal Module are specific to SpikeFlow designed to leverage SNN-specific functionalities. The Multiplexers facilitate sharing of crossbar columns with flash ADCs, Shift-and-Add circuit and the DIFF modules. The number of columns shared by an ADC, Shift-and-Add and DIFF module depends on the \textit{MUX size} parameter shown in Table \ref{tab:spikesim_parameters}. Shift-and-Add circuit are incorporated to support bit-splitting of weights. Typically, crossbars in ANNs \cite{chen2018neurosim} require separate Shift-and-Add circuit to support input bit-serialization and bit-splitting for weights. Due to the binary spike input, SpikeFlow only requires shift-and-add circuit to support weight splitting and not input serialization. Additionally, binary spike data allows replacing the traditional dual crossbar approach \cite{chen2018neurosim} for performing signed MAC computation with fully digital DIFF modules which reduce SpikeFlow's crossbar area significantly compared to ANN crossbars. Note, the crossbar area in case of SpikeFlow includes the area of the IMC device area, Input Peripherals, Multiplexers, ADCs, Shift-and-Add circuit and DIFF modules. The Input Peripherals contain switch matrices \cite{chen2018neurosim} for selecting the bit-lines (BL0-BLX), select lines (SL0-SLX) and word lines (WL0-WLX). The BLs are used to provide input $V_{read}$ voltage to the IMC device terminal. WLs facilitate crossbar row selection and SLs carry the output current in each column. In case of RRAM IMC-crossbars, the input peripherals additionally contain level-shifters that provide higher write voltages for RRAM device programming \cite{chen2018neurosim}.

\begin{figure}[h!]
    \centering
    \includegraphics[width=1\linewidth]{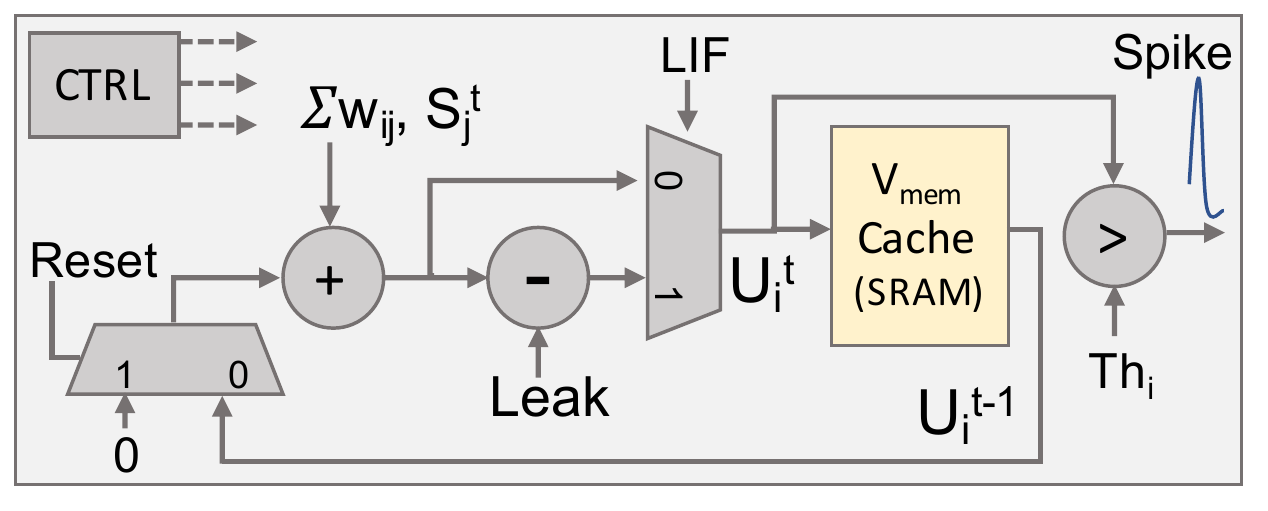}

    \caption{Figure showing the architecture of the LIF/IF Neuronal Module.}
    \label{fig:neuron_module}
     
\end{figure}
\textbf{LIF/IF Neuronal Module} In the LIF/IF Neuronal module shown in Fig. \ref{fig:neuron_module}, the MAC outputs from the GA at time-step $t$ and layer $i$, $\sum w_{ij} S_j^t$ are summed with the membrane potential $U^{t-1}_i$ (membrane potential at time $t-1$ fetched from the $V_{mem}$ cache) or $0$ (if $t=0$ or membrane potential is reset because of neuron firing). Depending on the $LIF$ signal being high or low, the output of the adder (integration functionality) or the subtractor (leaky-integration functionality) is passed on to the comparator. A neuron spikes at time $t$ (produces a binary `1') if the  membrane potential $U_i^t$ is greater than the threshold value ($Th_i$ which is specific to layer $i$). The neuronal spike output is relayed via NoC to the Tile Input Buffer of the tile mapping the preceding layer. The destination address of the spike is determined by the address appended with the spike. Note, that the $V_{mem}$ Cache is an SRAM memory required to store the membrane potentials over multiple time-steps. As we will see in Section \ref{sec:neuron_ar}, $V_{mem}$ has a significant contribution to the overall hardware overhead in SpikeFlow. 

\begin{figure}[h]
    \centering
    \includegraphics[width=\linewidth]{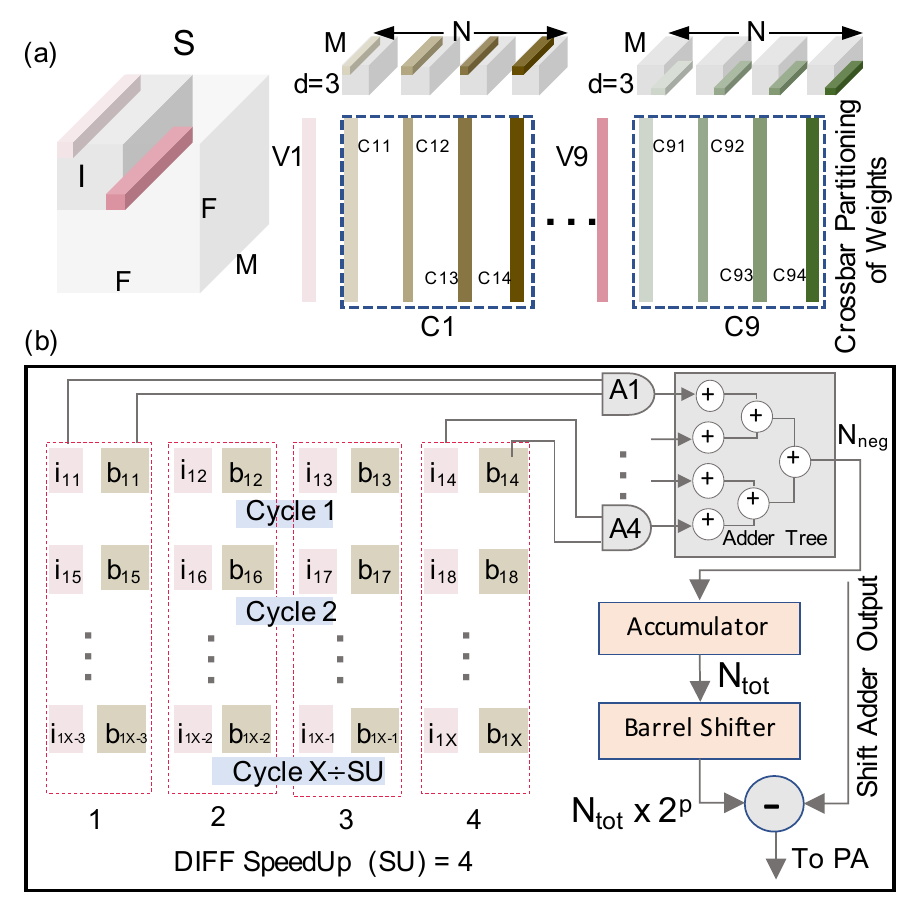}

    \caption{(a) Illustration of SNN kernel and input partitioning for mapping onto the SpikeFlow architecture. (b) The architecture and computation flow of the {DIFF} Module.}
    \label{fig:diff_module}
     
\end{figure}

\textbf{Mapping SNNs onto SpikeFlow:} SpikeSim employs the weight mapping strategy used in Neurosim \cite{chen2018neurosim} for partitioning and mapping the pre-trained software SNN weights. Fig. \ref{fig:diff_module}a, shows how a weight kernel of size N$\times$M$\times$d$\times$d is partitioned on X$\times$X crossbars. Here, $N$ and $M$ are the output and input channel dimensions, respectively and $d$ is the kernel dimension. The weight kernels are partitioned along the input channel ($M$) dimension and mapped along a crossbar column. While, the corresponding weights along the output channel dimension ($N$) are mapped on different columns of the same crossbar. The weights along the $d\times d$ dimensions are mapped on different crossbars. Hence, for a $d$= 3 kernel, 9 crossbars (C1-C9) are required. A similar partitioning is applied to the spike input map $S$ over the M$\times$d$\times$d section. This mapping strategy maximizes input data reuse and minimizes the buffer access. The software weights are converted to IMC device conductance values using a linear mapping scheme \cite{chen2018neurosim, jain2020rxnn} (See section \ref{sec:NICE} and Fig. \ref{fig:weight_g_map} for more details). To understand how the SNN is mapped across crossbars, PEs, and Tiles in the SpikeFlow architecture, let us consider a scenario in which an SNN consists of three convolutional layers in succession with input ($M$)/output ($N$) channels as follows: 64/64, 64/128 and 128/512, kernel size $d$= 3, and circuit parameters $X$= 64, $N_{PE}$= 8 and $N_{C}$= 9. In this case, layer-I requires $(64\div64)*(64\div64)*3^2 = 9$ crossbars which require 1 PE to be mapped (or $PE_1 = 1$). In SpikeFlow, we adopt a design choice that an SNN layer can be mapped over multiple Tiles but multiple layers cannot be mapped in one Tile. Therefore, during mapping of layer I, the PEs are replicated over $N_{PE}$ processing elements resulting in parallel mapping. The parallelization $Par_i$ for layer $i$ is computed by $N_{PE} \div PE_i$. In case of layer I, the parallel mapping $Par_1 = N_{PE} \div PE_1= 8$. Similarly, the weights of layers II and III require on $PE_2 = 2$ and $PE_3 = 16$. Consequently, $Par_2=4$ and $Par_3=1$. In total, a total of 4 Tiles are required for mapping the 3-layered SNN.


\textbf{The DIFF module: } In general, a dual-crossbar approach \cite{truong2014new} is employed to implement signed-MAC operations for ANNs. However, this approach is highly hardware-intensive, requiring additional energy expended on analog computations in the crossbars as well as the energy-hungry ADCs. Furthermore, with the dual-crossbar approach, the total hardware area overhead of the PEs get doubled. An interesting by-product of the non-ideality-aware weight encoding (section \ref{sec:NICE}) is that the dual-crossbar approach can be replaced with a lightweight, fully digital DIFF module shown in Fig. \ref{fig:diff_module}b. 

The DIFF module services the dot-product outputs from the Shift-and-Add circuit inside a Crossbar and performs accurate signed-MAC operations. Before the DIFF computations begin, the sign bits of weights and the spike inputs to a Crossbar are loaded from the PE Input Buffer into flip-flops $b_{cj}$ and $i_{cj}$, respectively as shown in Fig. \ref{fig:diff_module}b. The sign bits and spike inputs correspond to the $j^{th}$ row of the $c^{th}$ crossbar. $b_{cj}$ stores `1' for a negative-valued weight and a binary `0' for a positive-valued weight. For each crossbar column, the DIFF module requires $X\div SU$ cycles where, $SU$ denotes the {DIFF SpeedUp} value. DIFF SpeedUp dictates the number of AND gates and the width of the adder tree for parallely computing multiple rows of a column. For example, in case of a crossbar with size $X$= 64, a DIFF with $SU$= 4 will complete the operations in $X\div SU= 16$ cycles for one column.

In each cycle, the spike input $i_{cj}$ and the corresponding $b_{cj}$ undergo AND operation followed by addition in the Adder Tree. The Adder Tree outputs the count of negative weights receiving a spike input in each cycle $N_{neg}$. The $N_{neg}$ is accumulated over all the cycles resulting in $N_{tot}$. $N_{tot}$ represents the total count of negative weights in a column receiving spike inputs ($N_{tot}$). $N_{tot}$ scaled by $2^p$ (see Section \ref{sec:NICE} for the details on $p$) is subtracted from the Shift-and-Add circuit output to obtain the signed-MAC output for the column. The scaling is performed by a $p$-bit left shift operation using the Barrel Shifter.  

\begin{figure}[h!]
 \centering

\includegraphics[width=0.7\columnwidth]{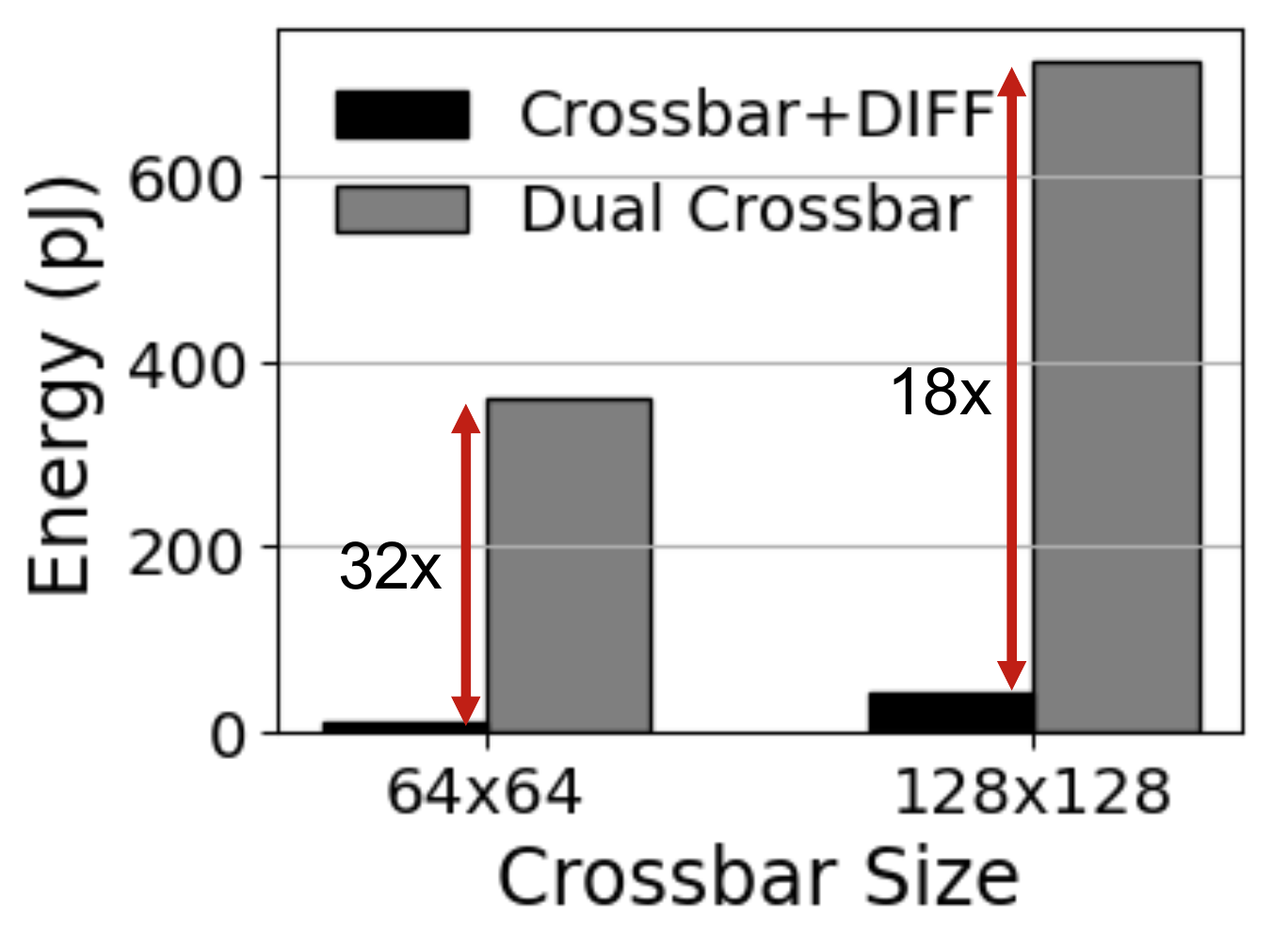}
  
\caption{Plot showing energy reduction achieved in case of DIFF module compared to the dual-crossbar approach for signed-MAC operations across $64\times64$ and $128\times128$ sized crossbars.}
\label{energy_red_diff}
\vspace{-2mm}
\end{figure}

Fig. \ref{energy_red_diff} shows that in our SpikeFlow architecture, the DIFF module results in $32\times$ and $18\times$ more energy-efficient MAC operations as opposed to the standard dual-crossbar approach for a single $64\times64$ and $128\times128$ crossbar-arrays, respectively. Here, we consider the total energy expended by the crossbar-array, input peripherals, multiplexer, ADC and the shift-adders.\\ 


\textbf{Computation Cycle:}
In SpikeFlow, crossbars are the fundamental MAC operation modules. First, the SL switch matrix converts the digital spike inputs to voltages ($V_{read}$ for spike inputs and $0V$ for no spike case). Simultaneously, the WL switch matrix activates WL0-WLX, enabling parallel read from all the IMC devices in the crossbar. The input voltages and programmed IMC device conductance values undergo multiplication and accumulation by virtue of Ohm's law and Kirchoff's current law, respectively leading to crossbar column currents along the bit-lines (BL0-BLX). The column currents are converted to digital partial sum values using flash ADCs.

The ADC outputs from each crossbar are fed to the Shift-and-Add circuit and successively to the DIFF module for obtaining the signed MAC output value. The DIFF module output is accumulated in the PA and stored in the PB. The outputs from PB are relayed to Tile Accumulator (TA). Post accumulation, the outputs are relayed to TB. Every entry in the TB is appended with the respective destination tile's address that maps the preceding layer. Next, the address appended values are pushed into the NoC router buffer which relays the TB outputs to the global accumulator. The outputs from GA are fed directly to the LIF/IF Neuronal module for the non-linear activation. Post activation, the spike data is transmitted over the NoC to the destination Tile.

\subsection{NICE: Non-ideality Computation Engine}
\label{sec:NICE}

    

\begin{algorithm}[t]
    \caption{NICE's Evaluation Flow}

\RaggedRight \textbf{Input}: Pre-trained SNN with $N_l$ layers, weights $W$, and $T$ time-steps, Crossbar Size $X$, ADC Precision ($h$), $R_{off}$, $R_{on}$, Bits/Cell, Variation $\sigma$, Wire resistance $r$\\
  
\textbf{Output} Hardware-Realistic Accuracy: 

  \begin{algorithmic}[1]\\
    \State{\textcolor{violet}{/* Before SNN Inference */}}
    \For{$i = 1$ to $N_l$}
        \If{``Conv" or ``Linear" layer}
            \State{$p$ = ceil($log_2(min(|W_{ideal, i}|))$)}
            \State{$W_{Enc}$ = $W_{ideal, i}$} \Comment{Initialization} 
            \For{$w_j \in W_{Enc}$}
                \If{$w_j$ $<$ 0}
                    \State{$w_j$ = $w_j + 2^p$} \Comment{\begin{tabular}{l}
                 NI-aware\\ Encoding
            \end{tabular}} 
                \Else \State{$w_j$ = $w_j$}
                \EndIf
            \EndFor
            \State{$W_{Enc, ~Noisy}$ = $W_{Enc} + \mathcal{N}(0,\sigma)$} \Comment{\begin{tabular}{l}
                 Device\\ Variations
            \end{tabular}}\\
            \State{Partition $W_{Enc, ~Noisy}$ into crossbars of size $X$.}
            \State{Create coefficient matrix $A_i$.}
        \EndIf
    \EndFor\\
        \State{\textcolor{violet}{/* During SNN Inference */}}
        
        \For{$t$ = 1 to $T$}
            \For{$i$ = 1 to $N_l$}
                \If{``Conv" or ``Linear" layer}
                    \State{Create $B_{t,i}$ using the layer $i$'s inputs at $t$.}
                    \State{$I$ $\gets$ solution of system $A_iI$ = $B_{t,i}$}
                    \State{$MAC_{u}$ $\gets$ $I$ converted to $h$-bit digital value.}
                    \State{\textcolor{OliveGreen}{/* Inside DIFF Module */}}
                    \State{Compute $N_{tot}$ for each crossbar column}
                    \State{$MAC_{s}$ = $MAC_{u} - N_{tot} \times 2^p$}
                \EndIf
                
            \EndFor
        \EndFor
    
  \end{algorithmic}
       \label{algorithm:nice}
    
\end{algorithm}
Our \textit{Non-ideality Computation Engine} (NICE) is designed in Python to compute the accuracy of SNNs mapped onto the SpikeFlow architecture. This computation engine captures the impact of hardware-level resistive crossbar non-idealities to generate MAC outputs during inference. 

\textbf{Need for the NICE simulator:} Prior works for ANNs, such as Neurosim \cite{chen2018neurosim} and CrossSim \cite{plimpton2016crosssim} that carry out crossbar-based inference of ANNs, neglect the inclusion of resistive parasitic non-idealities. A recent framework called {RxNN} \cite{jain2020rxnn}, although takes into account resistive non-idealities, requires long simulation time to generate non-ideal ANN weights from the ideal ones and performs input-independent modelling of non-idealities. To this effect, {GenieX} \cite{chakraborty2020geniex} captures input-dependent non-idealities for ANN inference, but follows an empirical approach by training an auxiliary fully-connected network to model the crossbar non-idealities. This makes transferability of GenieX across different crossbar sizes difficult as the model requires re-training for every new crossbar size. To this end, NICE adopts a generalized circuit analysis-based method that incorporates input data dependency and makes it transferable across different SNN architectures and crossbar array sizes.
\begin{figure}
    \centering
    
   \includegraphics[width=
   \columnwidth]{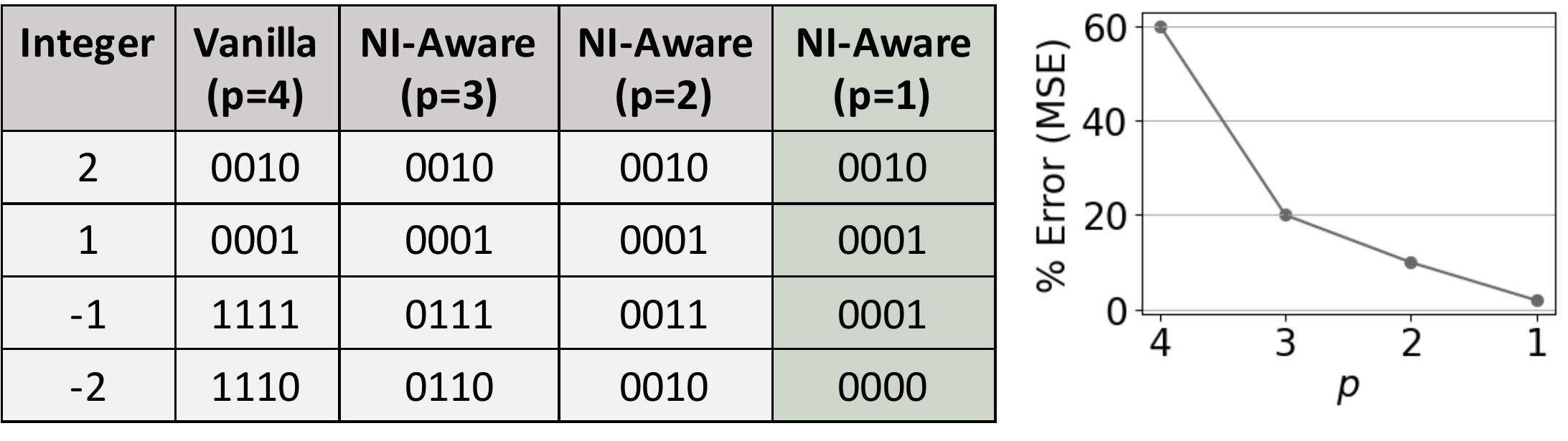} 
    \caption{(left) Table showing an example of NI-aware weight encoding. (b) The percentage mean squared error (MSE) between software (Pytorch) and hardware-realistic (NICE) MAC outputs for different $p$ values.}
    \label{fig:NICE_example}
\end{figure}
\begin{figure}
    \centering
    
   \includegraphics[width=
   0.5\textwidth]{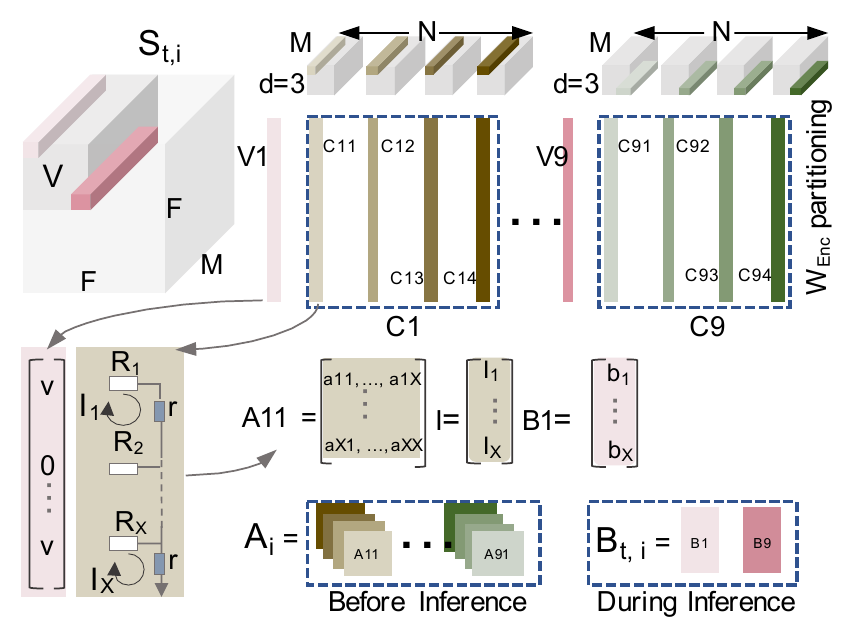} 
    \caption{Figure showing a demonstration of NICE's computation flow.}
    \label{fig:NICE_demo}
\end{figure}

\textbf{Non-ideality-aware Weight Encoding}
As shown in Algorithm \ref{algorithm:nice}, all pre-trained weights of the convolution and linear layers undergo non-ideality aware (NI-aware) encoding. For each layer $i$, we first initialize $W_{Enc}$ with software SNN weights $W_{ideal, i}$. The NI-aware encoding encodes the negative weights to unsigned values thereby, increasing the number of $0$s. This results in higher resistance values of the IMC devices mitigating the non-ideal effects of interconnect parasitics in crossbars \cite{bhattacharjee2021switchx, bhattacharjee2021neat}. The value of $p$ is specific to layer $i$'s weight distribution. Fig. \ref{fig:NICE_example}(left) illustrates an example of a 4-bit ($k=4$) quantized layer having weights $W=$ \{-2, -1, 1, 2\} in its distribution. For a 4-bit quantization, $p$ = 4 represents the naive 2's complement representation of the weights. As $p$ reduces, the number of 0's in the representation of -1 and -2 increases. For the given weight distribution, the maximum number of $0$s are obtained for $p = ceil(log_2(min(|W|))) = ceil(log_2(|-2|))$ \textit{i.e,} at $p$ = 1. Here, the weights -1 and -2 are encoded as 1 and 0, respectively which increases the number of $0$s compared to the vanilla case and hence results in higher IMC device resistances. Earlier works \cite{jain2020rxnn, royfundamental} have shown that higher device resistances ensure less effect of crossbar non-idealities in analog MAC computations and hence lower mean squared error (MSE) between software and hardware MAC outputs (Fig. \ref{fig:NICE_example}(right)). 

Now, we describe the computation flow of NICE as follows as has been illustrated in Fig. \ref{fig:NICE_demo}.

\begin{wrapfigure}{l}{0.4\columnwidth}
 \centering
\includegraphics[width=0.4\columnwidth]{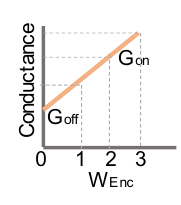}
\caption{Figure illustrating linear mapping of NI-aware weights $W_{Enc}$ to IMC device conductances.}
\label{fig:weight_g_map}
\vspace{-2mm}
\end{wrapfigure}

\textbf{Creating Coefficient Matrix ($A_i$): } After the NI-aware encoding, device conductance variation with distribution $\mathcal{N}(0,\sigma)$ is added to the encoded weights $W_{Enc}$ to create $W_{Enc, ~Noisy}$. Next, $W_{Enc, ~Noisy}$ is partitioned into analog crossbars of size $X$ (See Section \ref{sec:spikeflow} for details). Next, each crossbar column is converted to resistance ladders of size $X$. Depending on the weight value, the device conductance values (or resistances $R_1$-$R_X$) are linearly mapped as shown in Fig. \ref{fig:weight_g_map}. To model the IR-drop non-ideality, we incorporate column resistances $r$ at each device node. For each resistance ladder a coefficient matrix $Axy$ ($x^{th}$ column of the $y^{th}$ crossbar) is generated. $A_i$ is the stack of all the coefficient matrices $Axy$ for layer $i$. Additionally, $A_i$ is statically computed before the SNN inference begins.

\textbf{Generating $B_{t,i}$ during SNN Inference} During SNN inference, the input spike map at time $t$ and layer $i$, $S_{t,i}$ is partitioned according to the crossbar size and sent to the rows of the crossbar $Vp$ (for the $p^{th}$ crossbar). The $Vp$ matrix contains the input voltage values ($V_{read}$= $v$ and $0$ represents spike and no-spike, respectively). The $Vp$ matrix is used to create the matrix $B_{t,i}$. The output column currents ($I_X$) of all the crossbars are obtained by solving the system of linear equation $A_iI$ = $B_{t,i}$. Here, $I$ represents the variable matrix of the linear equation system. The output column currents $I$ are converted to $h-bit$ digital values where, $h$ is the ADC precision. 

\textbf{Obtaining Signed MAC outputs ($MAC_s)$): }Due to NI-aware encoding, the resulting MAC values obtained after ADC quantization are unsigned ($MAC_u$). To convert the unsigned outputs to signed values $MAC_{s}$, the DIFF module computes $N_{tot}$ for each crossbar column. As mentioned in Section \ref{sec:spikeflow}, $N_{tot}$ is the number of negative weights in each crossbar column receiving a spike input. $N_{tot}$, scaled by $2^p$ ($p$ determined during NI-aware encoding) is then subtracted from $MAC_u$ resulting in $MAC_s$. Multiplying the factor $2^p$ compensates for the $2^p$ factor added to the negative weights during NI-aware weight encoding. Note, that the given method of NI-aware weight encoding feasibly works due to the binary nature of spike inputs. Due to this, the value of $N_{tot}$ is merely the number of negative weights that received spike inputs. In case when there are multi-bit inputs, the calculation of $S$ is not straight-forward. 


\begin{figure*}
    \centering
    \includegraphics[width=
       0.75\textwidth]{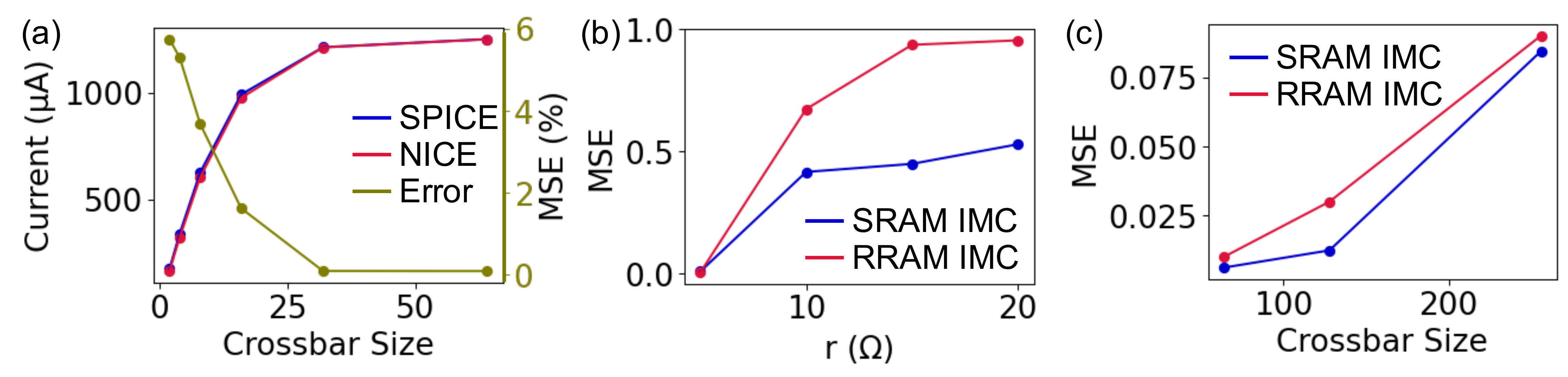}
    \caption{Figure showing the (a) column currents for different crossbar sizes obtained using NICE and SPICE simulations and percentage MSE values. The primary y-axis denotes the crossbar column current ($\mu$A) and the secondary y-axis denotes the percentage MSE of NICE MAC outputs with respect to SPICE counterparts (b) MSE between software (Pytorch) and hardware-realistic (NICE) MAC outputs for different $r$ values (c) MSE between software and hardware MAC outputs for different crossbar sizes and IMC devices- SRAM \cite{jaiswal20198t} and RRAM \cite{hajri2019rram}. }
    \label{fig:NICE_analysis}
\end{figure*}

Fig. \ref{fig:NICE_analysis}a shows that the crossbar currents obtained using a Cadence Virtuoso SPICE simulator and our NICE engine have very close resemblance with very small percentage error for crossbar sizes above 32$\times$32. Here, the value of interconnect parasitic resistance is taken as $5\Omega$. Thus, NICE computes circuit-accurate currents that incorporate the impact of crossbar non-idealities during inference. Fig. \ref{fig:NICE_analysis}b and Fig. \ref{fig:NICE_analysis}c further corroborate the efficacy of NICE in capturing the crossbar non-ideality. As expected, the \textit{Mean-squared-error} (MSE) between the hardware and software convolution operations increase as the bit-line resistances $r$ and crossbar sizes increase. This has also been observed in prior literature \cite{chakraborty2020geniex}. 





\subsection{ELA Engine}
\label{sec:ela_engine}
Our (ELA) {Engine} is designed in Python to compute the hardware-realistic energy, latency and area for a SpikeFlow-mapped SNN model. 

\textbf{Inference latency evaluation:} The latency of a particular layer $i$ (convolution or linear layer) has two components as shown in Eq. \ref{eq:tot_latency}. First, the $Tile ~Latency_i$ required for performing all the computations inside the Tile and second, the $NoC ~Latency_i$ for performing all the communications in the particular layer. 
\begin{equation}
    Latency_i = Tile ~Latency_i + NoC ~Latency_i
    \label{eq:tot_latency}
\end{equation}
$Tile ~Latency_i$ can be computed using Eq. \ref{eq:lat_tile}. Here, $Cycles_{i}$ denotes the number of clock cycles required for all the operations in the layer $i$ and $Clock ~Period$ denotes the period of one clock cycle. 

\begin{equation}
    Tile ~Latency_i = Cycles_{i} \times Clock ~Period
    \label{eq:lat_tile}
\end{equation}

$Cycles_i$ can be computed using Eq. \ref{eq:cycles} where, $N_{ops,i}$ denotes the number of operations (convolutions or linear operations) per output channel in layer $i$ and ${Cycles/Op}$ is the number of clock cycles required for one convolution operation. In crossbar architectures, multiple PEs in a Tile execute parallely to perform one or more operation. Therefore, $Cycles/Op$ depends on the PE latency ($\alpha$) and the parallel mapping in layer $i$ ($Par_i$) as shown in Eq. \ref{eq:c_per_conv}. 

\begin{equation}
    Cycles_{i} = {Cycles}/ {Op} \times N_{ops, i}
    \label{eq:cycles}
\end{equation}
\begin{equation}
    {Cycles/ Op} = \frac{\alpha}{Par_i}
    \label{eq:c_per_conv}
\end{equation}

The PE Latency ($\alpha$) is defined as the number of clock cycles required for reception of spike-inputs by PEs, MAC computations inside the PEs by all $N_C$ crossbars, digital subtractions using the DIFF module, accumulation in PA and finally storage of the MAC outputs in PB. $\alpha$ has a fixed value depending on the number of crossbars inside PE ($N_C$) and the SpeedUp of the DIFF module. 

\begin{figure}[h!]
    \centering
    \includegraphics[width=\columnwidth]{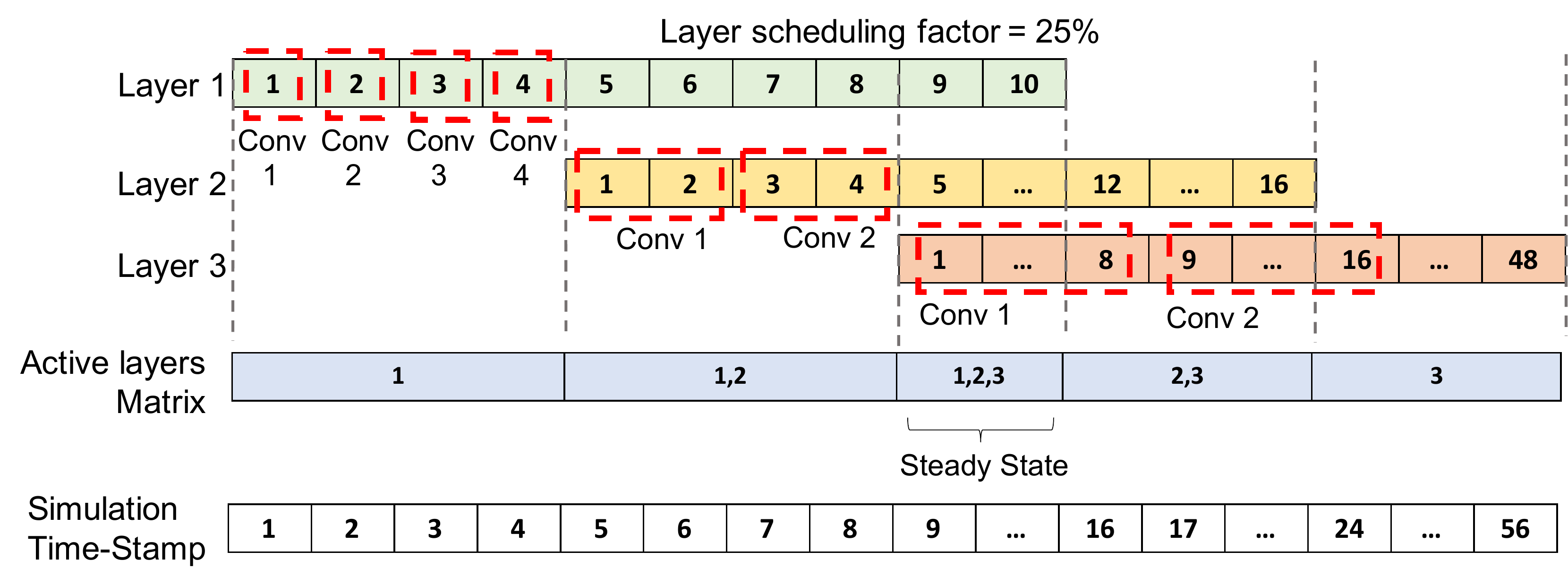}
    \caption{Figure illustrating an example of SpikeFlow's pipelined mode of inference.}
    \label{fig:layer_scheduling}
\end{figure}
\textbf{Pipelined Mode of Inference: } The SpikeFlow-mapped SNN is executed in a pipelined dataflow. We define a \textit{layer-scheduling factor} which determines the percentage of total operations in layer $i$ that must be completed before layer $i+1$ computations begin. Fig. \ref{fig:layer_scheduling} shows the pipelined execution of the 3-layered SNN discussed in Section \ref{sec:spikeflow}. For illustration, let layers 1, 2 and 3 require 10, 8 and 6 convolution operations per output channel ($N_{conv,1}$= 10, $N_{conv,2}$= 8 and $N_{conv,3}$= 6), respectively. Additionally, we will assume that the PE latency $\alpha=8$ and the SNN requires only one time-step for this illustration. Note, that the SNN's time-step is different from the Simulation Time-Stamps shown in Fig. \ref{fig:layer_scheduling}. Based on the values of $Par_1$, $Par_2$, and $Par_3$, layers 1, 2 and 3 require $\frac{\alpha}{8}$, $\frac{\alpha}{4}$ and $\frac{\alpha}{1}$ cycles per convolution, respectively. Due to layer scheduling of 25\%, layer 2 begins after 0.25$\times N_{conv,1}=$ 4 convolutions of layer 1 and layer 3 begins after 2 convolutions of layer 2. The trace generator in the ELA engine tracks the number of active layers over all the simulation time stamps.

The trace generator further records the \textit{Steady State} during the SNN execution. Steady State is defined as the state at which the maximum number of SNN layers are active. The ELA engine estimates the optimal $V_{mem}$ cache size required for the LIF/IF neuronal module using the steady state information. As we will see in Section \ref{sec:neuron_ar}, the steady state directly impacts the hardware area consumed by the LIF/IF Neuronal module.

Next, we add the NoC communication latency to the tile latency using the approach proposed in \cite{krishnan2021siam} to obtain the final hardware latency. The NoC latency is computed separately using a cycle-accurate NoC simulator \cite{jiang2013detailed, krishnan2021siam}. The NoC latency for layer $i$ depends on the latency of single packet transfer ($Packet ~Latency$) and number of data packets ($N_{p,i}$) transmitted between the tiles mapping layer $i$ and destination tiles mapping layer $i+1$ (see Eq. \ref{eq:NoC_lat}). The number of data packets is computed using Eq. \ref{eq:Np}. Here, $A_i$ denotes the number of activations in layer $i$, $k_{mem}$ is the quantization of membrane potential value $U$ and $NoC ~width$ denotes the NoC channel width.

\begin{equation}
NoC ~latency_i = N_{p,i} \times Packet ~Latency
\label{eq:NoC_lat}
\end{equation}
\begin{equation}
N_{p,i} = \frac{A_i\times k_{mem}}{NoC ~width}
\label{eq:Np}
\end{equation}

\textbf{Inference energy \& area evaluations:} The hardware energy per inference and area estimations by the ELA Engine is straight-forward. For a given set of SNN, circuit and device parameters, after the SNN is mapped on the SpikeFlow architecture, the number of Tiles, PEs, crossbars are determined. Further, the trace generator estimates the optimal $V_{mem}$ size of the neuronal module. The total hardware area is the sum of the area occupied by the individual components. Likewise, as the inference progresses on the SpikeFlow architecture, the total dynamic hardware energy expended is the sum of the dynamic energy consumed by the individual active components at a given point in time. 


  
  
        

                
        
        
        
        
            
    
    

\section{Experiments and Results}
\begin{figure*}[t]
    \centering
    \includegraphics[width=0.8\linewidth]{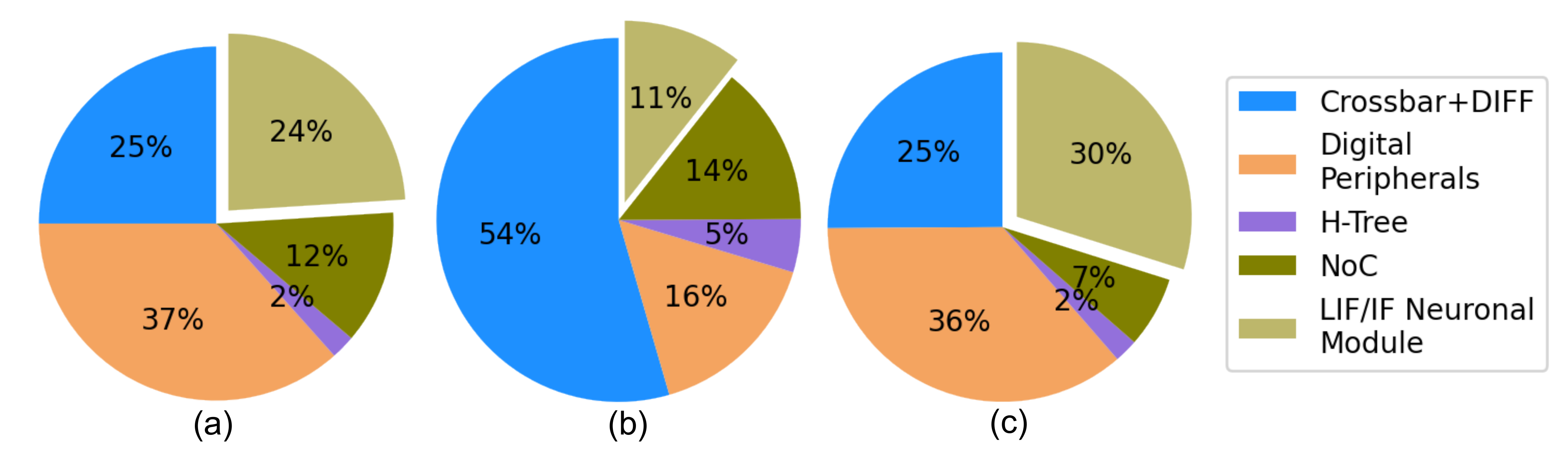}

    \caption{Pie-chart showing component-wise area distribution for (a) VGG9/CIFAR10 ($T$= 5) (b) VGG16/CIFAR100 ($T$= 10) and (c) VGG16/TinyImagenet ($T$= 10) SNNs. All SNNs are evaluated on the SpikeFlow architecture with 64$\times$64 crossbars. }
    \label{fig:pie_area}
     
\end{figure*}

\begin{figure*}[t]
    \centering
    \includegraphics[width=0.8\linewidth]{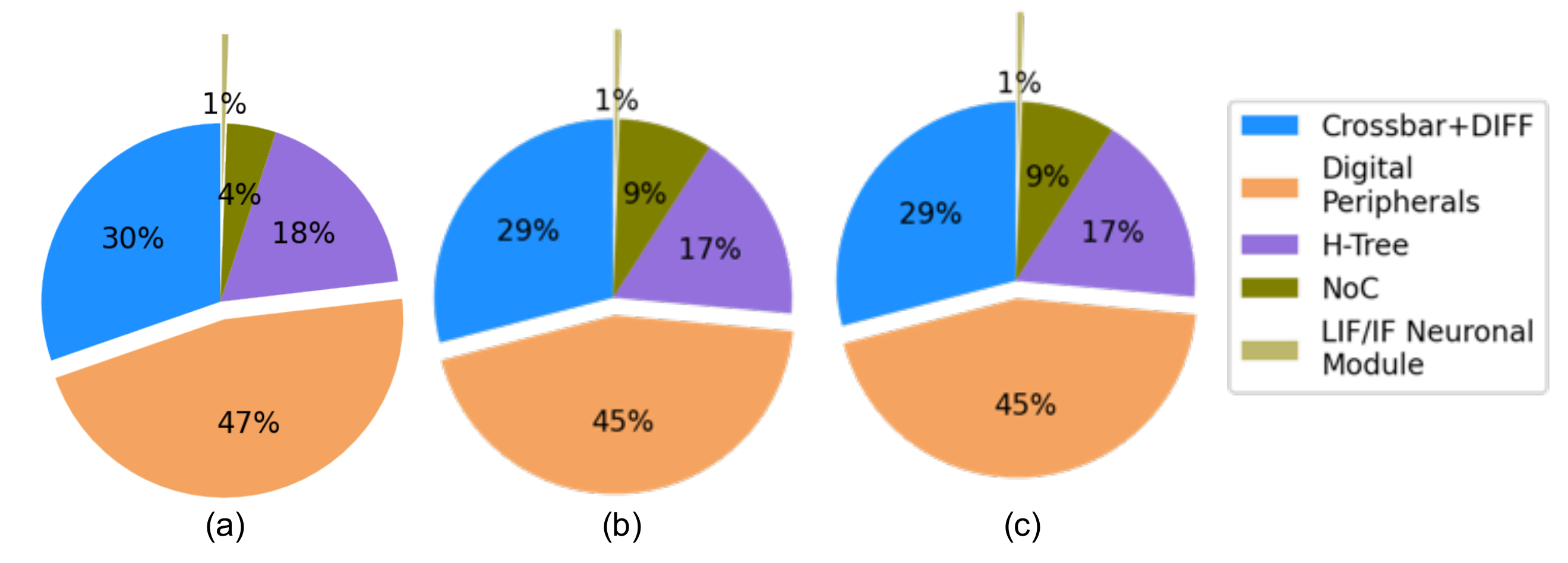}

    \caption{Pie-chart showing the component-wise inference energy expended by (a) VGG9/CIFAR10 ($T$= 5), (b) VGG16/CIFAR100 ($T$= 10) and VGG16/TinyImagenet ($T$= 10) SNNs when evaluated on a SpikeFlow architecture with 64$\times$64 crossbars. }
    \label{fig:pie_energy}
     
\end{figure*}
\subsection{Experimental Setup}

\label{sec:expt}

\begin{table}[h!]
    \centering
    \caption{Table showing SNN parameters used for the benchmark datasets.}
    \resizebox{1\columnwidth}{!}{
    \begin{tabular}{|l|l|c|c|l|c|}
    \hline
    \multirow{5}{*}{\rotatebox[origin=c]{90}{{\textbf{CIFAR-10}}}} & \multicolumn{2}{c|}{SNN Parameters} & \multirow{5}{*}{\rotatebox[origin=c]{90}{{\textbf{CIFAR-10}}}} & \multicolumn{2}{c|}{SNN Parameters} \\ 
    \cline{2-3} \cline{5-6}
    
    & {{Network Topology}} &  {VGG5} && {{Network Topology}} & {VGG9}\\ \cline{2-3} \cline{5-6}
    
    & {{$k$}} &  {4-bit, 1-bit} && {{$k$}} & {4-bit, 1-bit}\\ \cline{2-3} \cline{5-6}
    & {{$k_{mem}$}} &  {8-bits} && {{$k_{mem}$}} & {8-bits}\\ \cline{2-3} \cline{5-6}
    & {{$T$}} &  {5} && {{$T$}} & {5}\\ \cline{2-3} \cline{5-6}
    
    & {{Activation}} &  {LIF} && {{Activation}} & {LIF}\\ \hline
    
    \multirow{5}{*}{\rotatebox[origin=c]{90}{{\textbf{CIFAR-100}}}} & \multicolumn{2}{c|}{SNN Parameters} & \multirow{5}{*}{\rotatebox[origin=c]{90}{{\begin{tabular}{c}
    \textbf{Tiny}\\\textbf{Imagenet} 
    \end{tabular}}}} & \multicolumn{2}{c|}{SNN Parameters} \\ \cline{2-3} \cline{5-6}
    
    & {{Network Topology}} &  {VGG16} && {{Network Topology}} & {VGG16}\\ \cline{2-3} \cline{5-6}
    
    & {{$k$}} &  {4-bit, 1-bit} && {{$k$}} & {4-bit, 1-bit}\\ \cline{2-3} \cline{5-6}
    & {{$k_{mem}$}} &  {8-bits} && {{$k_{mem}$}} & {8-bits}\\ \cline{2-3} \cline{5-6}
    & {{$T$}} &  {10} && {{$T$}} & {10}\\ \cline{2-3} \cline{5-6}
    
    & {{Activation}} &  {LIF} && {{Activation}} & {LIF}\\ \hline
    \end{tabular}}
    
    \label{tab:SNN_params}
\end{table}
\textbf{Datasets: }In this work, we benchmark pretrained SNN model using our Python-based SpikeSim tool. The inference is carried out using benchmark datasets, namely CIFAR10, CIFAR100 and TinyImagenet. CIFAR10 and CIFAR100 datasets consist of RGB images (50,000 training and 10,000 testing) of size 32$\times$32 belonging to 10 and 100 classes, respectively. The TinyImagenet dataset is a more complex dataset with RGB images (100,000 training and 10,000 testing) of size 64$\times$64 belonging to 200 classes. 

\textbf{SpikeSim Parameters: } The SNN Parameters for different datasets have been provided in Table \ref{tab:SNN_params}. All the SNNs are trained using Back-Propagation Through Time (BPTT) algorithm using Adam optimizer with an initial learning rate of $1e-3$. For the BPTT training, the inputs are direct rate encoded \cite{kim2022rate}. We use BPTT as it achieves higher performance with fewer time-steps compared to standard ANN-SNN conversion methods \cite{sengupta2019going,diehl2015fast,han2020deep,li2021free,rueckauer2017conversion}. Although we use BPTT-based direct encoded models, SpikeSim can evaluate different pre-trained SNN models irrespective of the training algorithm used. Unless stated otherwise, the circuit and device-specific parameters used in the experiments with SpikeSim have been listed in Table \ref{tab:xbar_params}.


\begin{table}[h!]
    \centering
    \caption{Table enlisting the values of various Circuit and Device parameters used for experiments with SpikeSim.}
    \resizebox{0.6\columnwidth}{!}{
    \begin{tabular}{|l|c|} \hline
       \multicolumn{2}{|c|}{\textbf{Circuit Parameters}} \\ \hline
       NoC Topology & Mesh \\ \hline
       $SU$ & 64\\ \hline
       Sheduling Factor & 25\% \\ \hline
       Clock Frequency & 250 MHz \\ \hline 
       $N_C$ & 9 \\ \hline
       $N_{PE}$ & 8 \\ \hline
       $MUX ~Size$ & 8 \\ \hline
       $B_{GB}$, $B_{TB}$, $B_{PB}$ & 20KB, 10KB, 5KB \\ \hline
       $B_{TIB}$, $B_{PIB}$ & 50KB, 30KB \\ \hline
       NoC Width & 32bits \\ \hline
       $V_{DD}$ & 0.9V \\ \hline
       $V_{read}$ & 0.1V \\ \hline
       $h$ & 4-bits \\ \hline
       $r$ & 5$\Omega$ \\ \hline
       \multicolumn{2}{|c|}{\textbf{SRAM IMC Device \cite{jaiswal20198t} Parameters}} \\ \hline
       Technology & 65 nm CMOS \\ \hline
       Bits/Cell & 4 \\ \hline
       $R_{on}$ and $R_{off}$ &  416.67$\Omega$ and $\infty$\\ \hline
       $\sigma$ & 0.1 \\
       \hline
        \multicolumn{2}{|c|}{\textbf{RRAM IMC Device \cite{hajri2019rram} Parameters}} \\ \hline
       Bits/Cell & 1 \\ \hline
       $R_{on}$ and $R_{off}$ &  20k$\Omega$ and 200k$\Omega$\\ \hline
       $\sigma$ & 0.1 \\
       \hline
    \end{tabular}}
    
    \label{tab:xbar_params}
\end{table}

\subsection{Hardware Realistic Performance Evaluation using NICE}
\label{sec:nice_res}

\begin{figure}[h!]
    \centering
    \includegraphics[width=\columnwidth]{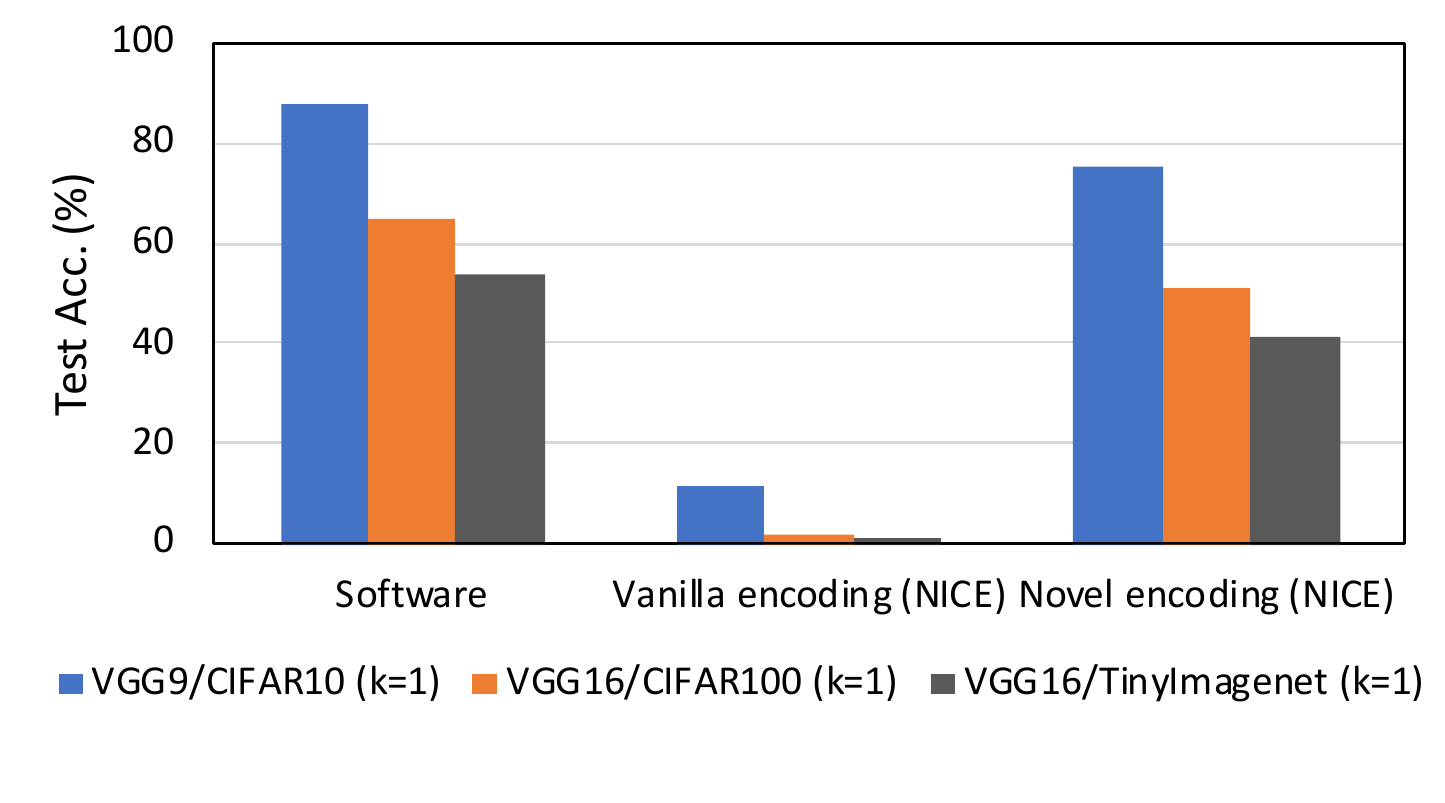}

    \caption{Bar chart showing the software and NICE evaluated SNN models using SpikeSim tool with RRAM IMC device (1-bit/Cell).}
    \label{fig:NICE_acc}
     
\end{figure}

Fig. \ref{fig:NICE_acc}, shows that the NI-aware weight encoding significantly improves the performance of crossbar-mapped SNNs compared to vanilla encoding. Vanilla encoding on SpikeSim leads to SNN accuracy degradation to random values of $\sim11\%$, $\sim2\%$ and $\sim1\%$ for CIFAR10, CIFAR100 and TinyImagenet datasets, respectively. This is attributed to error accumulation across multiple time-steps owing to resistive crossbar non-idealities \cite{bhattacharjee2022examining}. With our novel NI-aware weight encoding scheme, we find that the inference accuracy of VGG9/CIFAR10, VGG16/CIFAR100 and VGG16/TinyImagenet SNNs improve more than $70\%$, $50\%$ and $40\%$, respectively, compared to the vanilla encoding due to the reduction in crossbar non-ideal effects.


\subsection{Area and Energy Distributions}

\label{sec:ela_res}

Fig. \ref{fig:pie_area} and Fig. \ref{fig:pie_energy} shows the component-wise area and energy distribution for SNNs with VGG9/CIFAR10, VGG16/CIFAR100 and VGG16/TinyImagenet network topology. All results correspond to 64$\times$64 crossbar mapping with RRAM IMC device having 1-bit/Cell. Here, \textit{Crossbar} includes the IMC crossbar array, multiplexer, ADC, shift-adder and input peripherals. The \textit{digital peripherals} consist of PA, PB, PE Input Buffer, TA, TB, Tile Input Buffer, GB, GA and PO modules shown in Fig. \ref{fig:spikeflow}. It can be observed that the area and energy values increase with larger network sizes as deeper networks require more tiles and more computations. Furthermore, for the same SNN topology (VGG16), the neuronal module area and the total energy increase by $>6\times$ and $>4\times$ with increase in the input feature dimension from 32 (CIFAR100) to 64 (TinyImagenet). This is because of a higher number of convolution operations per SNN layer. 


\subsection{Neuronal Area Overhead and Mitigation}

\label{sec:neuron_ar}

From Fig. \ref{fig:pie_area}, we can observe that the total neuronal area accounts for a significant percentage share in the overall chip area. This share is as large as $24\%$ ($1.2 mm^2$) for the VGG9 SNN, which is almost equal the total chip area consumed by the PE and the DIFF units. For a VGG16 SNN inferred using a more complex dataset like TinyImagenet with input dimensions= 64, the absolute area encompassed by the neuronal unit becomes as large as $9.32 mm^2$. This observation is because the LIF/IF neuron module contains a large $V_{mem}$ SRAM cache to store the intermediate membrane potentials over multiple time steps (see Fig. \ref{fig:neuron_module}). Due to the pipelined dataflow, multiple layers can be active at the steady state and thus the LIF/IF $V_{mem}$ cache size must be large enough to store the membrane potentials of the active layer neurons. To this end, we propose mitigation strategies to ameliorate the LIF/IF neuron module's area overhead. 


\begin{figure}[h!]
    \centering
    \includegraphics[width=.8\columnwidth]{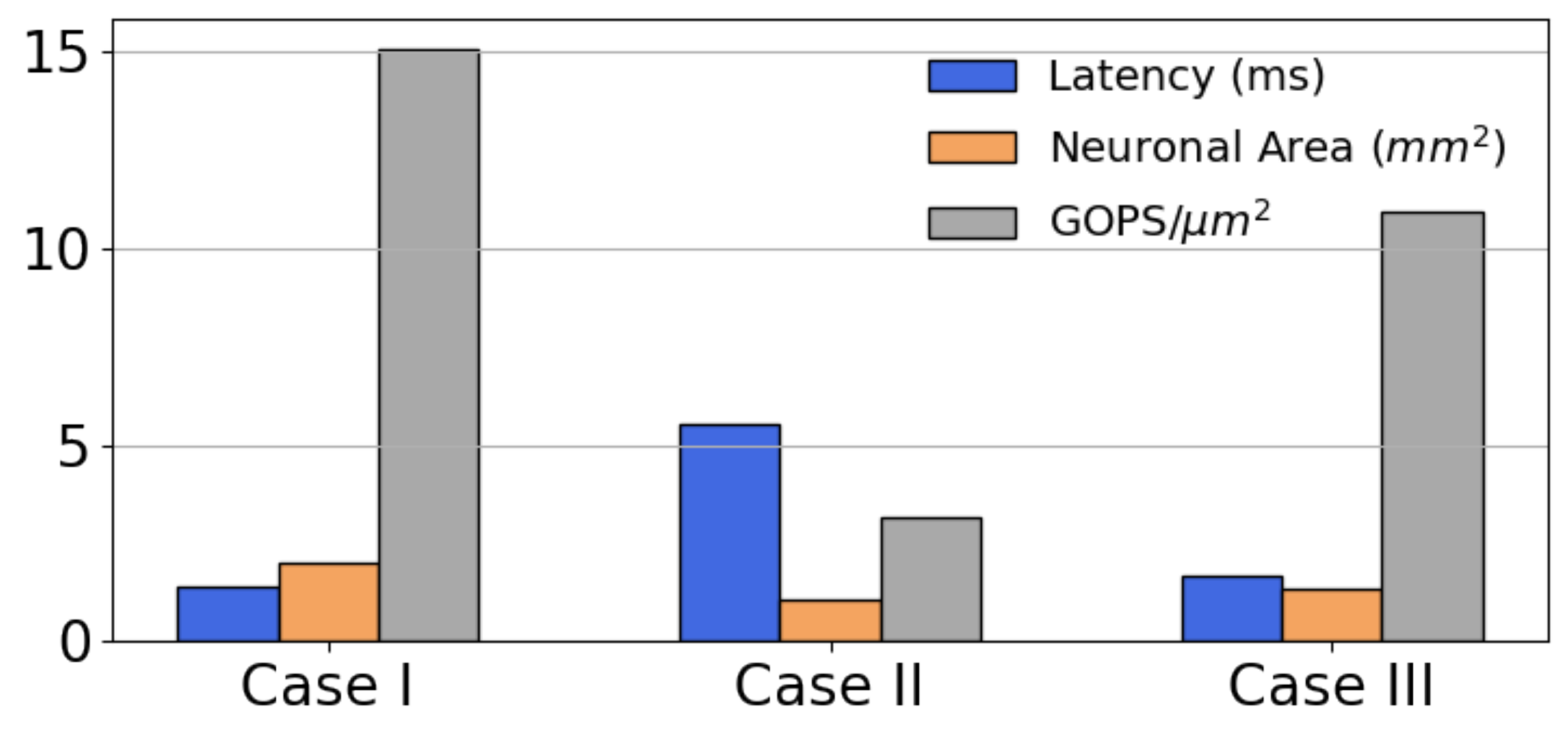}

    \caption{Bar-chart showing the trade-off between neuronal area, total inference latency and average $GOPS/\mu m^2$ in the steady state for a VGG9/CIFAR10 SNN inferred on a SpikeFlow for different dataflows. The y-axis denotes Latency (ns), Neuronal Area (mm$^2$), and average GOPS/$\mu m^2$ values.}
    \label{fig:various_scheduling}
     
\end{figure}
In Fig. \ref{fig:various_scheduling}, we analyze the trade-offs among three different dataflow strategies for the VGG9/CIFAR10 SNN mapped on a 64$\times$64 RRAM crossbar. Case-I refers to SpikeFlow's standard pipelined dataflow with scheduling factor (explained in Section \ref{sec:ela_engine}) of $25\%$ for each layer. The number of active layers in the steady state are 3. In Case-II, we analyse the effect of the tick-batched dataflow introduced in \cite{narayanan2020spinalflow}. In tick-batching, layers are processed sequentially and all the timesteps for a specific layer are computed together before proceeding to the next layer (scheduling factor = $100\%$). This leads to a $2\times$ reduction in the total neuronal area. However, tick-batching is highly impractical for crossbar-mapped SNNs as the inference latency increases $4\times$. Furthermore, the average value of $GOPS/\mu m^2$ in the steady state is reduced by $5\times$ due to significant hardware area-underutilization. In Case-III, we adopt a layer-specific scheduling dataflow wherein, we heuristically assign different scheduling factors for each convolutional layer to reduce the number of active layers in the steady state. Here, the typical scheduling factors for different layers range between 25\% and 75\%. Due to this, the number of active layers is reduced to 2. We find that the neuronal area reduces by $1.5\times$ with a marginal increase in inference latency by $1.17\times$ and reduction in the average $GOPS/\mu m^2$ by $1.38\times$. Although the layer-specific scheduling method reduces the neuronal area overhead it inevitably entails trade-offs between the neuronal area overhead,  inference latency and hardware underutilization. However, our objective is to minimize both neuronal area as well as inference latency. To this end, we propose SNN topological modifications guided by hardware evaluations using SpikeSim to achieve higher area and compute efficiency. 

\begin{figure}[h!]
    \centering
    \includegraphics[width=0.8\columnwidth]{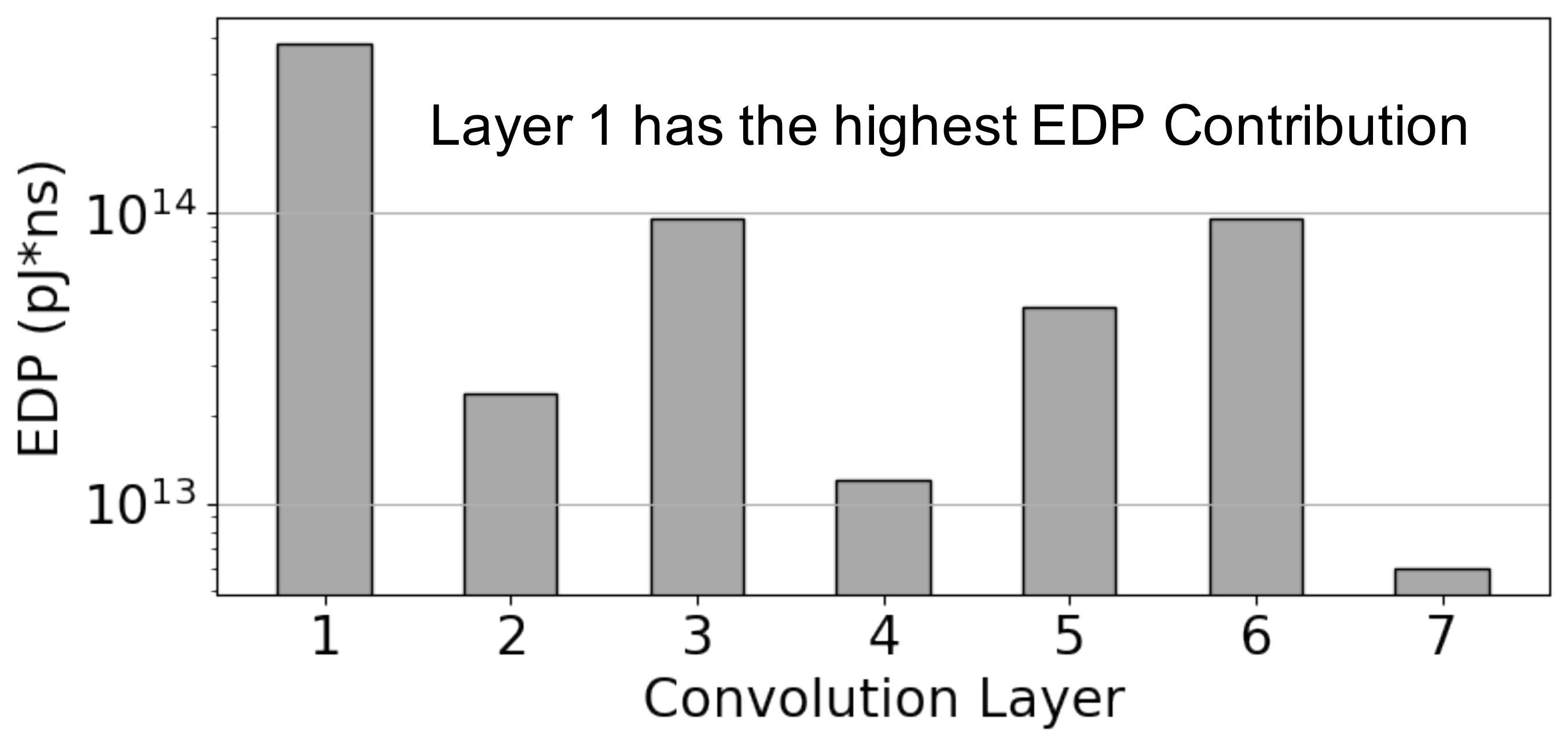}

    \caption{Plot showing the layerwise EDP on SpikeSim for the VGG9/CIFAR10 SNN topology with 5 time-steps. }
    \label{fig:layer_edp}
     
\end{figure}

\begin{figure}[h!]
    \centering
    \includegraphics[width=0.8\linewidth]{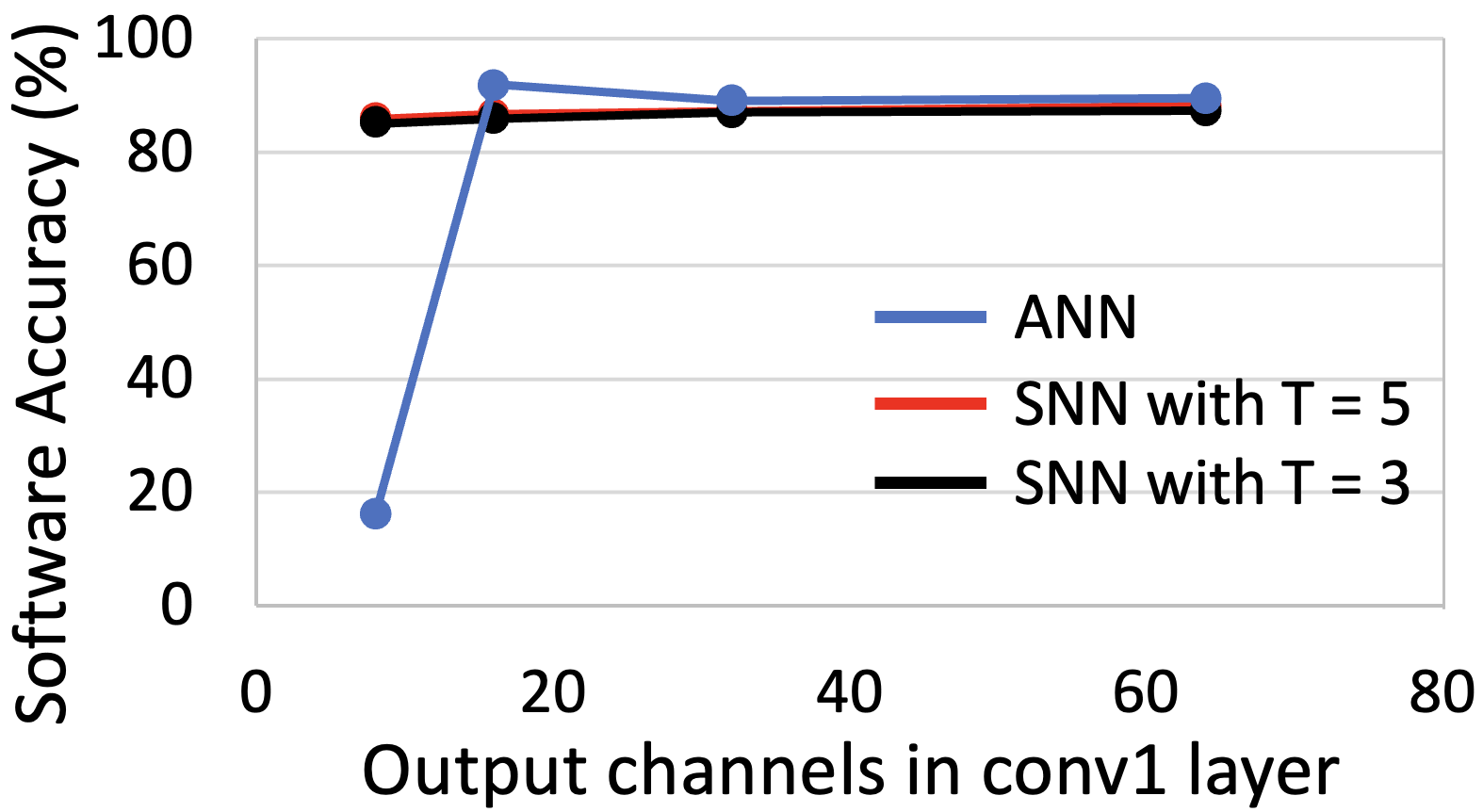}

    \caption{Plot showing the CIFAR10 software accuracies of a 4-bit VGG9 ANN and 4-bit VGG9 SNN  models with different channel scaling across different number of time-steps $T$.}
    \label{fig:ann_vs_snn_ts_acc}
     
\end{figure}

Fig. \ref{fig:layer_edp} presents how the total hardware Energy-Delay-Product (EDP) per inference varies across different convolutional layers of the VGG9/CIFAR10 SNN mapped on 64$\times$64 RRAM crossbar. We observe that the first convolution layer incurs the highest EDP in comparison to the other layers. This is because the first layer incurs higher convolution operations due to larger input feature dimension. Thus, our proposed topological modifications would be pertinent to the first convolutional (or conv1) layer. We adapt and train the VGG9/CIFAR10 SNN by reducing the number of output channels for the first layer from 64 to 8. Fig. \ref{fig:ann_vs_snn_ts_acc} shows the software accuracy of VGG9/CIFAR10 SNN and ANN with channel scaling in the first layer. The VGG9 ANN's software accuracy declines drastically to $\sim16\%$ on decreasing the number of output channels in conv1 layer to 8. In contrast, due to the additional time-step dimension, the accuracy of SNNs is only marginally reduced to 87\% with time-steps 3 and 5 facilitating channel scaling modifications. 

\begin{figure}[h!]
    \centering
    \includegraphics[width=0.9\columnwidth]{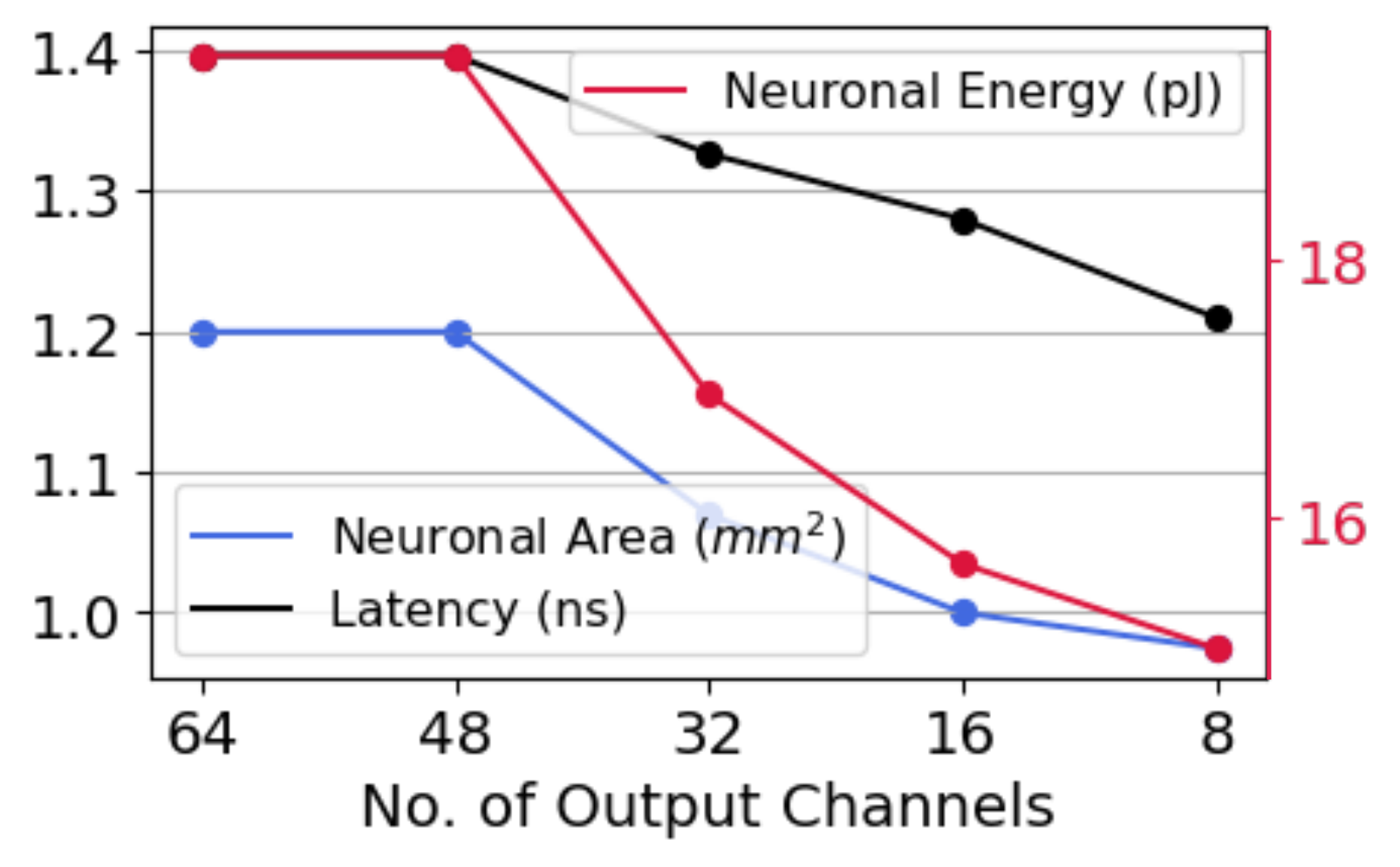}

    \caption{Trends for neuronal area, neuronal energy and inference latency with channel scaling on VGG9 SNN architecture evaluated on a SpikeFlow architecture with 64$\times$64 crossbars. The primary y-axis denotes Neuronal Area (mm$^2$)/ Latency (ns) and the secondary y-axis denotes Neuronal Energy (pJ).}
    \label{fig:after_topology_mod}
     
\end{figure}

\begin{figure}[h!]
    \centering
    \includegraphics[width=1\columnwidth]{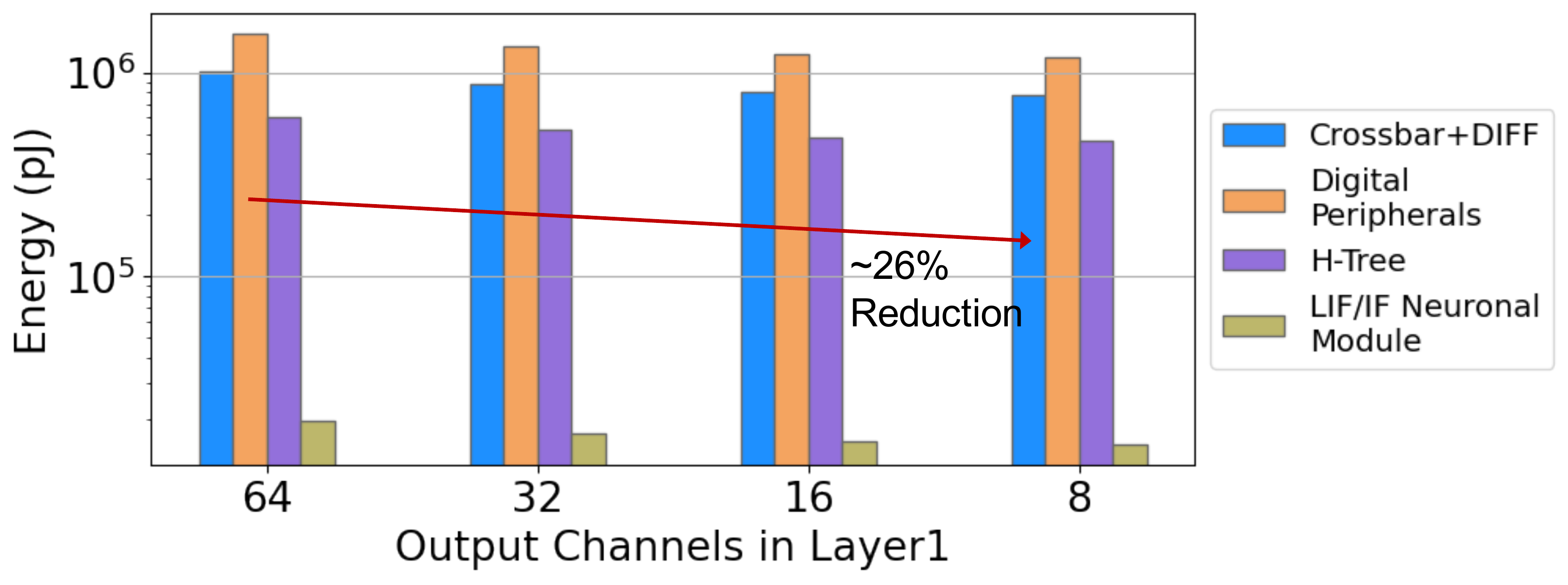}

    \caption{Bar-chart showing the variation in energies expended by different components in the SpikeFlow architecture (with 64$\times$64 crossbars) upon different channel scaling in the first layer of a VGG9 SNN. The energy reduction with channel-scaling is shown by the red arrow.}
    \label{fig:energy_periph_topology_mod}
     
\end{figure}

\begin{figure*}[t]
    \centering
    \resizebox{\textwidth}{!}{
    \begin{tabular}{ccc}
         \includegraphics[width=0.4\linewidth]{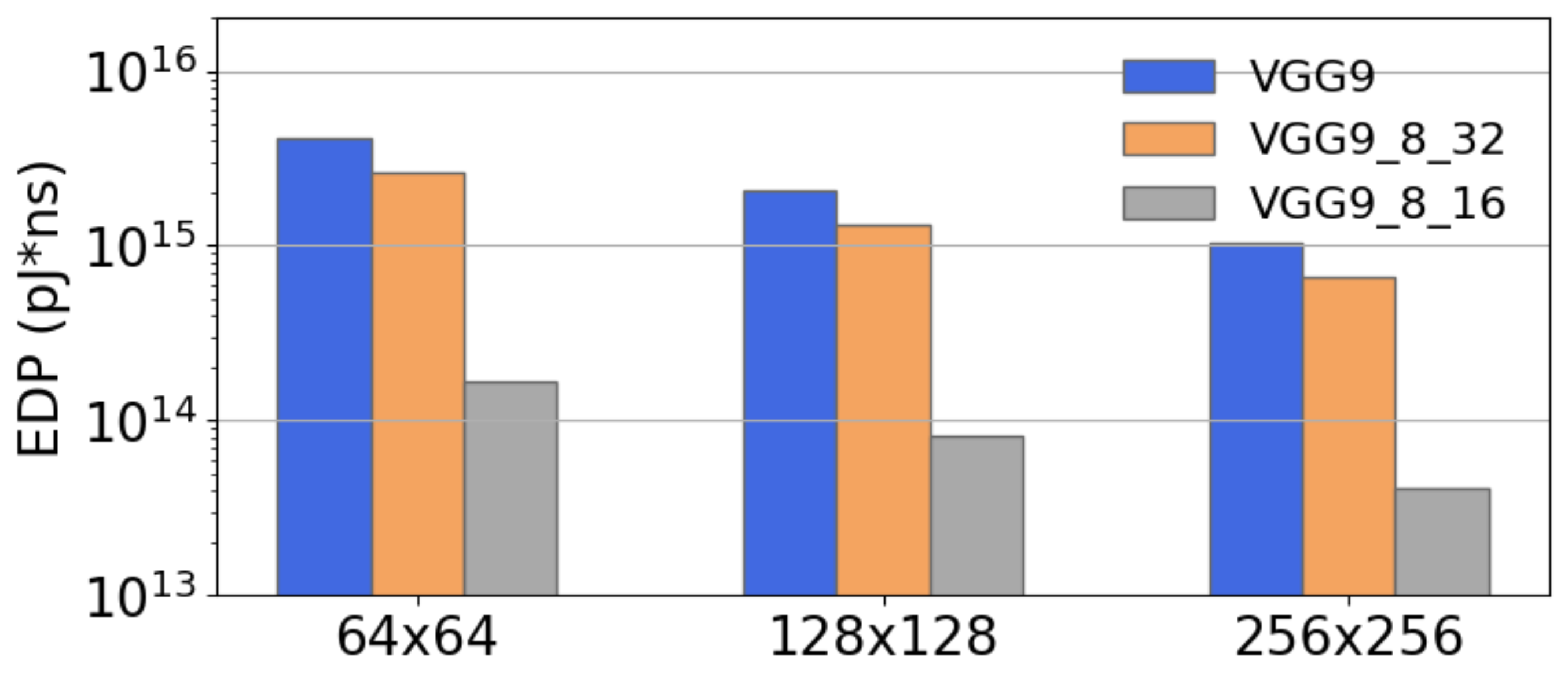} & \includegraphics[width=0.4\linewidth]{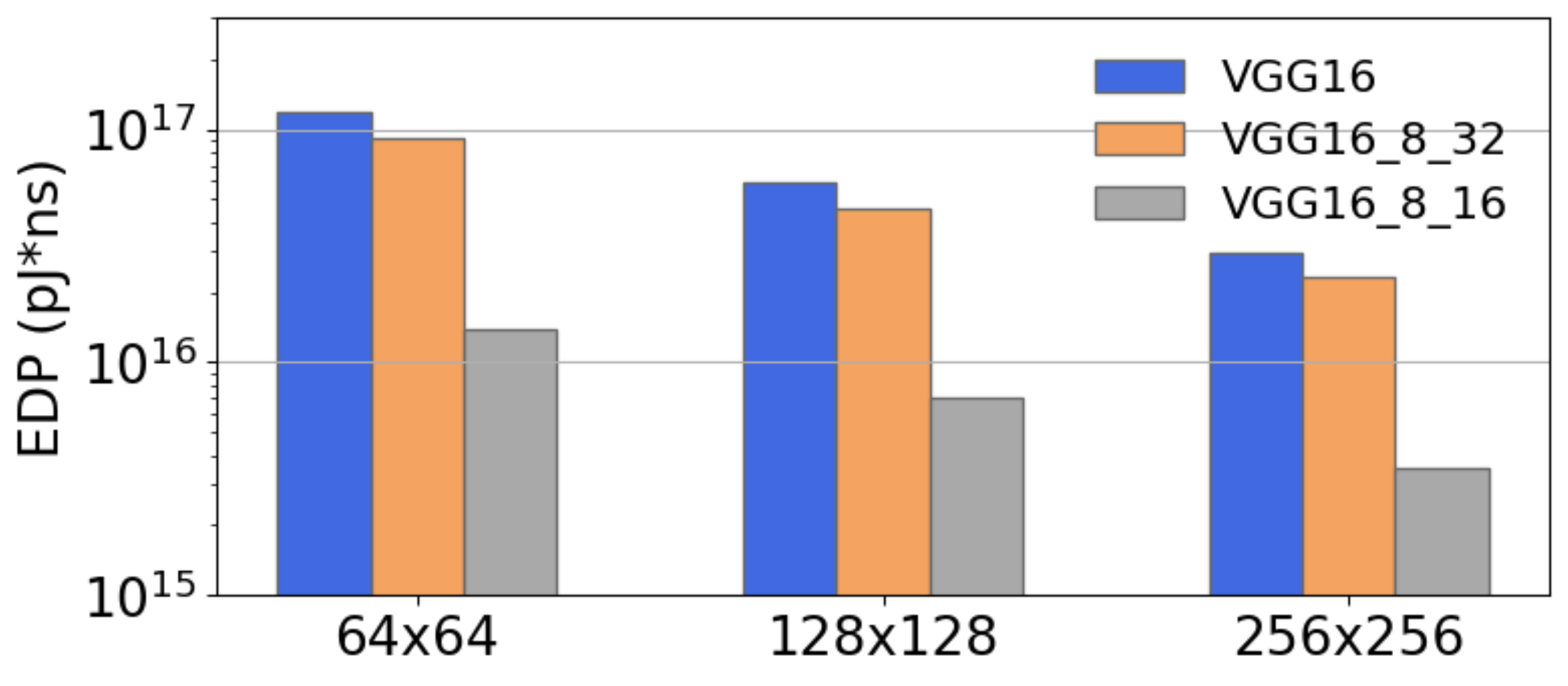}  & \includegraphics[width=0.4\linewidth]{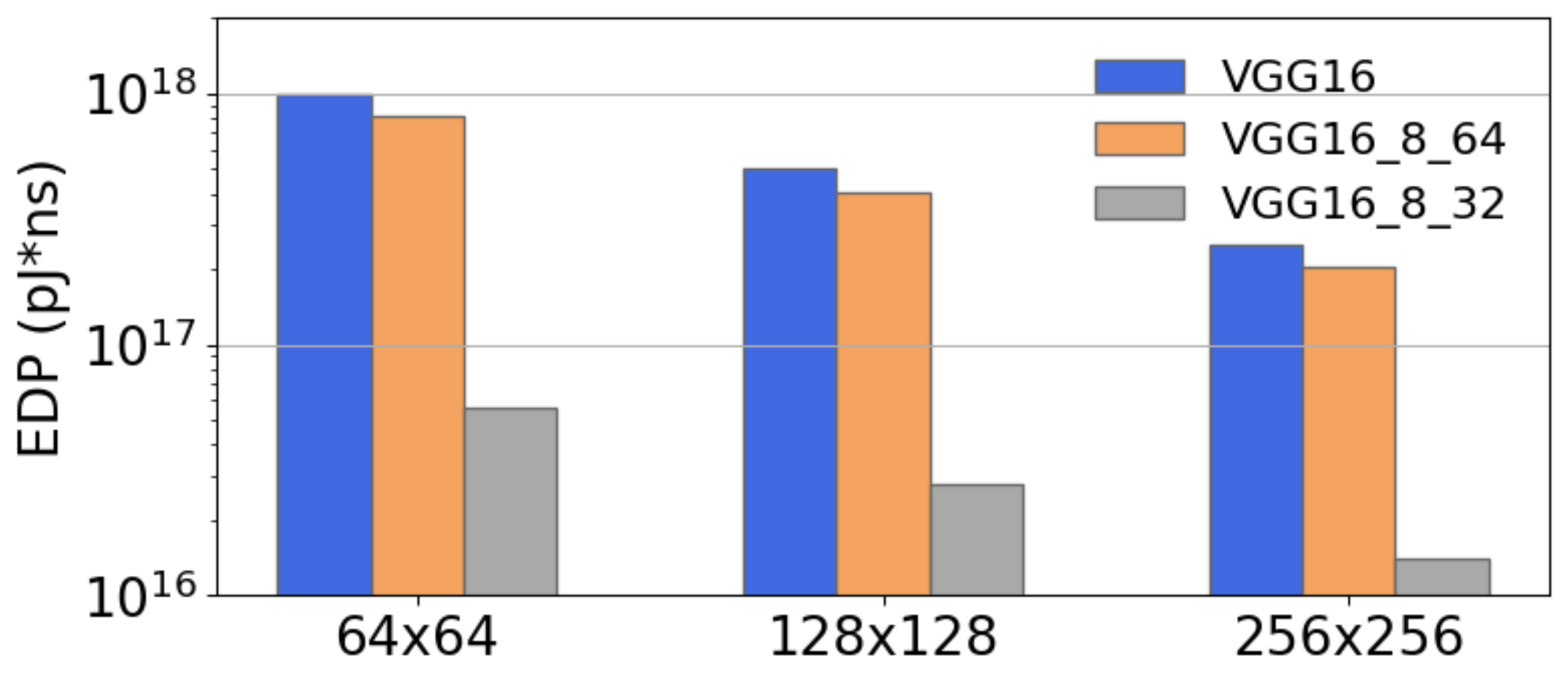} \\
         (a) & (b) & (c)\\

    \end{tabular}}

    \caption{Bar-chart showing EDP values in logarithmic scale for SNNs with standard- (a) VGG9/CIFAR10 ($T$=5), (b) VGG16/CIFAR100 ($T$=10), (c) VGG16/TinyImagenet ($T$=10) and models with channel and input dimension scaling across different crossbar sizes.}
    \label{fig:xbar_EDP_nets}
     
\end{figure*}

Fig. \ref{fig:after_topology_mod} presents the variation in neuronal area, neuronal energy and overall inference latency with respect to the number of output channels in the first layer. Overall with 8 output channels, the neuronal area overhead is reduced by $1.24\times$ along with a $1.15\times$ reduction in the overall inference latency. Consequently, the contribution of the neuronal area to the overall chip area is reduced to $20\%$ from $24\%$. Furthermore, due to channel scaling, we observe a $26\%$ reduction in the energy consumption of the neuronal module and other components as seen in Fig. \ref{fig:energy_periph_topology_mod}.


Fig. \ref{fig:xbar_EDP_nets} shows a holistic EDP evaluation for VGG9/CIFAR10 (Fig. \ref{fig:xbar_EDP_nets}a), VGG16/CIFAR100 (Fig. \ref{fig:xbar_EDP_nets}b) and VGG16/TinyImagenet (Fig. \ref{fig:xbar_EDP_nets}c) SNNs implemented on 64$\times$64, 128$\times$128 and 256$\times$256 analog crossbar arrays. Here, VGGX-N-M denotes a VGGX SNN model with N output channels in the conv1 layer and M$\times$M input feature dimensions. The results further corroborate that channel scaling in the first layer improves the energy efficiency of crossbar-mapped SNNs. Under similar SNN topology, mapping on a 256$\times$256-sized crossbar reduces the EDP by $4\times$ compared to a 64$\times$64 crossbar mapping. This is because, larger crossbars incur lower latency due to higher parallel MAC operations. Similarly, for the same crossbar size, input dimenstion scaling (16 from 32 in case of CIFAR10/CIFAR100 and 32 from 64 in case of TinyImagenet datasets),  reduces the EDP by $3.8\times$ compared to the model without any topological modifications. The highest EDP reductions ($\sim100\times$ compared to a standard SNN model on 64$\times$64 crossbar) is observed when a model undergoes channel and feature dimension scaling and mapped on the 256$\times$256 crossbar.

\section{Summary of Benchmarking Results using SpikeSim}
\begin{table*}
\caption{Table summarizing our SpikeSim based benchmarking results for the inference of various pretrained SNN topologies.}
\label{tab:summary}
\resizebox{\textwidth}{!}{%
\begin{tabular}{|c|c|c|cc|cc|cc|c|}
\hline
\multirow{2}{*}{\textbf{SNN   topology}} &
  \multirow{2}{*}{\textbf{Time-steps (T)}} &
  \multirow{2}{*}{\textbf{S/W accuracy   (\%)}} &
  \multicolumn{2}{c|}{\textbf{Accuracy using   NICE (\%)}} &
  \multicolumn{2}{c|}{\textbf{Energy ($\mu J$)}} &
  \multicolumn{2}{c|}{\textbf{Area ($mm^2$)}} &
  \multirow{2}{*}{\textbf{Latency ($ms$)}} \\ \cline{4-9}
 &
   &
   &
  \multicolumn{1}{c|}{\textbf{SRAM (4-bits)}} &
  \textbf{RRAM (1-bit)} &
  \multicolumn{1}{c|}{\textbf{SRAM (4-bits)}} &
  \textbf{RRAM (1-bit)} &
  \multicolumn{1}{c|}{\textbf{SRAM (4-bits)}} &
  \textbf{RRAM (1-bit)} &
   \\ \hline
\textbf{VGG5/CIFAR10}        & 5  & 84.55 & \multicolumn{1}{c|}{76.31} & 69.65 & \multicolumn{1}{c|}{10.5} & 10.1 & \multicolumn{1}{c|}{3.08} & 2.92 & 93.0 \\ \hline
\textbf{VGG9/CIFAR10}        & 5  & 88.11 & \multicolumn{1}{c|}{79.61} & 75.19 & \multicolumn{1}{c|}{16.9} & 16.1 & \multicolumn{1}{c|}{5.37} & 5 & \num{2.6e2} \\ \hline
\textbf{VGG16/CIFAR100}      & 10 & 65.23 & \multicolumn{1}{c|}{55.56} & 50.81 & \multicolumn{1}{c|}{71} & 69.9 & \multicolumn{1}{c|}{16.68} & 14.36 & \num{1.6e3}\\ \hline
\textbf{VGG9-8-16/CIFAR10}   & 5  &  85.87 & \multicolumn{1}{c|}{78.12} & 74.9 &  \multicolumn{1}{c|}{15} & 13.9 & \multicolumn{1}{c|}{4.87} & 4.5 & 41.5\\ \hline
 \textbf{VGG16-8-16/CIFAR100} & 10 & 64.39 & \multicolumn{1}{c|}{56.12} & 50.74 & \multicolumn{1}{c|}{65.2} & 62 & \multicolumn{1}{c|}{16.11} & 13.79 & \num{4.3e2} \\ \hline
 \textbf{VGG16/TinyImagenet} & 10 & 54.03 & \multicolumn{1}{c|}{45.24} & 41.11 & \multicolumn{1}{c|}{129.25} & 127.26 & \multicolumn{1}{c|}{33.51} & 31.2 & \num{6.4e3} \\ \hline
\end{tabular}%
}
\end{table*}
In Table \ref{tab:summary}, we provide holistic inference and hardware performance results using the SpikeSim tool for different SNN models trained on the CIFAR10 and CIFAR100 datasets. The results are shown for mappings on 64$\times$64 crossbars. Firstly, the SNN inference accuracy upon crossbar-mapping reduces from its software inference accuracy by $\sim9\%$ and $\sim13\%$ for 4-bit SRAM and 1-bit RRAM IMC units, respectively. It must be recalled that due to the inherent vulnerability of SNNs towards crossbar non-idealities \cite{bhattacharjee2022examining}, the inference accuracy of SNNs on SpikeSim (without non-ideality aware weight encoding) drops down to $\sim10\%$ from the corresponding software accuracy. Due to NICE's novel weight encoding method, the classification performance of SNNs is restored to a suitable range as shown in Table \ref{tab:summary}. Additionally, SNNs with topological modifications and input dimension scaling (VGGX-8-16) achieve $1.15\times$ lower energy, $1.03-1.1\times$ lower area, and $3.72-6.27\times$ lower latency compared to the standard SNN models.

\section{Comparison Between ANN and SNN}

\begin{figure*}[h!]
    \centering
    \includegraphics[width=0.75\textwidth]{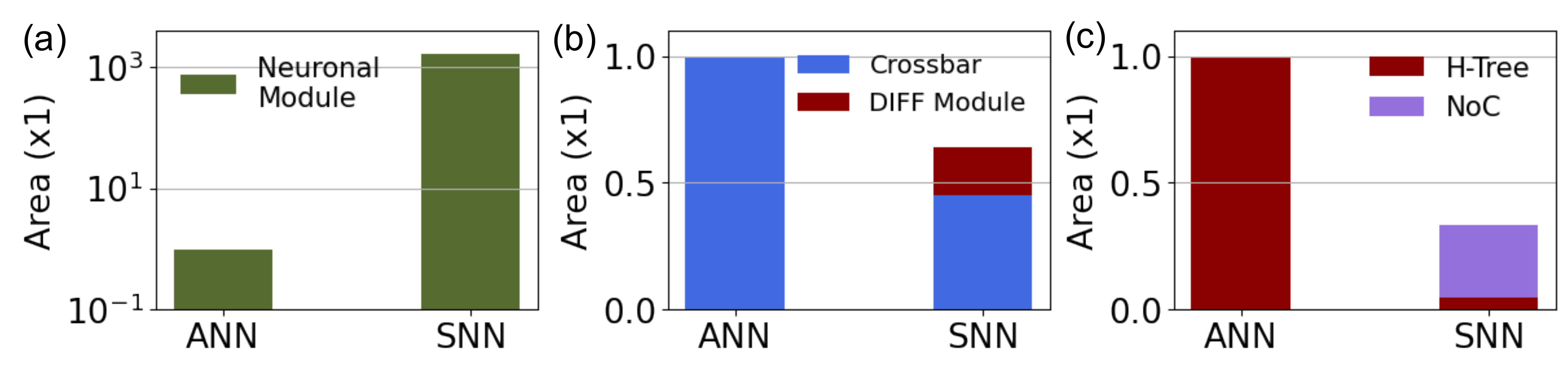}
    \caption{Figure showing the comparison for (a) Neuronal Module area (b) Crossbar Area and (c) Data Communication Area in Neurosim-based ANN and SpikeSim-based SNN implementations.}
    \label{fig:ANN_SNN_component_comp}
\end{figure*}

\begin{figure*}[h!]
    \centering
    \includegraphics[width=0.75\textwidth]{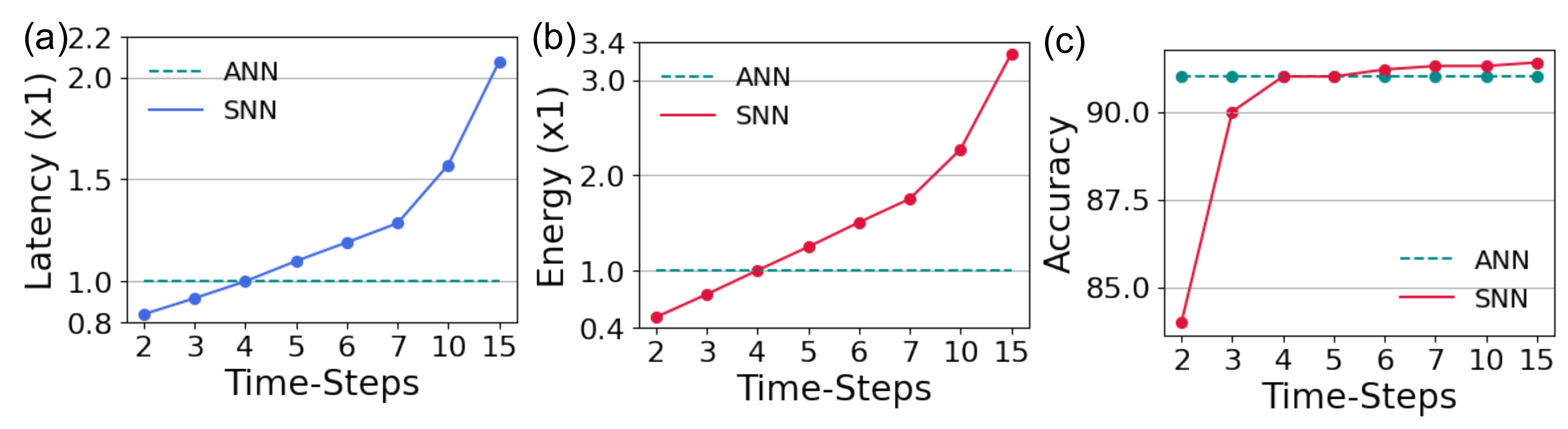}
    \caption{Figure comparing the (a) latency (b) energy and (c) software accuracy for an SNN across different time-steps. The latency and energy correspond to 4-bit SNN and are normalized with respect to a 4-bit quantized ANN baseline.}
    \label{fig:ANN_SNN_energy_lat_comp}
\end{figure*}

In Fig. \ref{fig:ANN_SNN_component_comp}, we perform a component-wise area comparison between a crossbar-mapped VGG9/CIFAR10 ANN and SNN. The ANN is mapped on the Neurosim \cite{chen2018neurosim} platform while the SNN is mapped on the SpikeSim platform. Both the mappings are based on 65nm CMOS technology on a 64$\times$64 SRAM crossbar array. Both the SNN and ANN are quantized to 4-bit precision. In addition, no channel or input dimension scaling is applied to the SNNs. Note, since the ANN and SNN is implemented on two different platforms all results are normalized with respect to Neurosim ANN  implementation to achieve a fair comparison.

As seen in Fig. \ref{fig:ANN_SNN_component_comp}a, the neuronal module's area for an SNN (the LIF/IF Neuronal Module) is around $\sim1000\times$ compared to ANN's ReLU Neuronal Module. This is because of the large SRAM cache to store membrane potentials ($V_{mem}$) across different time-steps. Fig. \ref{fig:ANN_SNN_component_comp}b shows that the crossbar area required by ANN is $1.8\times$ higher compared to SNNs. This is primarily because in SpikeSim, we replace the dual-crossbar approach used in ANNs for signed MAC operations with the DIFF module. Further, due to the binary spike data, SNN crossbars do not require Shift-and-Add circuits for input bit-serialization, and therefore consume 10\% lesser area compared to ANN crossbars. Finally, Neurosim uses H-Tree for all data communications both local and global to the tiles. In contrast, SpikeSim employs NoC and H-Tree interconnects for inter-tile and intra-tile communications, respectively. This reduces the data communication area in SpikeSim by $>3\times$ compared to ANNs as seen in Fig. \ref{fig:ANN_SNN_component_comp}c.  

From Fig. \ref{fig:ANN_SNN_energy_lat_comp}, we find that at sufficiently lower time-steps (T=3,4), SNNs can achieve iso-performance while being $1.3\times$ energy-efficient and $1.1\times$ faster compared to a 4-bit quantized ANN. This is because ANN crossbar architectures adopt input bit-serialization to encode multi-bit activation requiring multiple crossbar read cycles and hence higher computation and communication overheads. In contrast, SNNs require one read cycle over multiple time-steps to perform the MAC operations. If the number of time-steps are low enough, SNNs achieve higher compute efficiency. 

Therefore, lower number of time-steps will provide better advantage to SNNs in terms of energy and latency compared to ANNs. The current state-of-the-art SNN training methods require larger number of time-steps in order to achieve higher performance especially for complex datasets such as CIFAR100 and TinyImagenet. It is therefore critical for future research works to focus on achieving higher performance at lower time-step overhead for SNNs to be practically energy efficient.




\section{Conclusion}
This work presents SpikeSim, an end-to-end evaluation tool for hardware realistic inference performance (accuracy), energy, latency and area of crossbar-mapped SNNs. One of the important findings of this work involves- high area overhead of the neuronal module in SNNs due to temporal storage of membrane potentials for LIF/IF neuron functionality. To this end, we employ SNN topological modifications such as channel scaling, input scaling and layer-specific scheduling strategies to overcome the neuronal module's area. Based on our study, we recommend the following design strategies for future SNN-IMC-architecture co-design research- 1) Performing hardware-aware SNN architecture design to achieve low neuronal overhead and high crossbar utilization. 2) Designing IMC-architecture-specific dataflows to reduce number of active layers and thereby reduce the neuronal area overhead. 3) Exploiting the spatio-temporal properties in SNNs to achieve higher performance at lower time-steps in order to achieve superior energy efficiency and throughput compared to ANNs. 


\bibliographystyle{IEEEtran}
\bibliography{egbib.bib}

\begin{thebibliography}{10}
\providecommand{\url}[1]{#1}
\csname url@samestyle\endcsname
\providecommand{\newblock}{\relax}
\providecommand{\bibinfo}[2]{#2}
\providecommand{\BIBentrySTDinterwordspacing}{\spaceskip=0pt\relax}
\providecommand{\BIBentryALTinterwordstretchfactor}{4}
\providecommand{\BIBentryALTinterwordspacing}{\spaceskip=\fontdimen2\font plus
\BIBentryALTinterwordstretchfactor\fontdimen3\font minus
  \fontdimen4\font\relax}
\providecommand{\BIBforeignlanguage}[2]{{%
\expandafter\ifx\csname l@#1\endcsname\relax
\typeout{** WARNING: IEEEtran.bst: No hyphenation pattern has been}%
\typeout{** loaded for the language `#1'. Using the pattern for}%
\typeout{** the default language instead.}%
\else
\language=\csname l@#1\endcsname
\fi
#2}}
\providecommand{\BIBdecl}{\relax}
\BIBdecl

\bibitem{roy2019towards}
K.~Roy, A.~Jaiswal, and P.~Panda, ``Towards spike-based machine intelligence
  with neuromorphic computing,'' \emph{Nature}, vol. 575, no. 7784, pp.
  607--617, 2019.

\bibitem{stromatias2017event}
E.~Stromatias, M.~Soto, T.~Serrano-Gotarredona, and B.~Linares-Barranco, ``An
  event-driven classifier for spiking neural networks fed with synthetic or
  dynamic vision sensor data,'' \emph{Frontiers in neuroscience}, vol.~11, p.
  350, 2017.

\bibitem{kim2021optimizing}
Y.~Kim and P.~Panda, ``Optimizing deeper spiking neural networks for dynamic
  vision sensing,'' \emph{Neural Networks}, vol. 144, pp. 686--698, 2021.

\bibitem{davies2018loihi}
M.~Davies, N.~Srinivasa, T.-H. Lin, G.~Chinya, Y.~Cao, S.~H. Choday, G.~Dimou,
  P.~Joshi, N.~Imam, S.~Jain \emph{et~al.}, ``Loihi: A neuromorphic manycore
  processor with on-chip learning,'' \emph{Ieee Micro}, vol.~38, no.~1, pp.
  82--99, 2018.

\bibitem{akopyan2015truenorth}
F.~Akopyan, J.~Sawada, A.~Cassidy, R.~Alvarez-Icaza, J.~Arthur, P.~Merolla,
  N.~Imam, Y.~Nakamura, P.~Datta, G.-J. Nam \emph{et~al.}, ``Truenorth: Design
  and tool flow of a 65 mw 1 million neuron programmable neurosynaptic chip,''
  \emph{IEEE transactions on computer-aided design of integrated circuits and
  systems}, vol.~34, no.~10, pp. 1537--1557, 2015.

\bibitem{narayanan2020spinalflow}
S.~Narayanan, K.~Taht, R.~Balasubramonian, E.~Giacomin, and P.-E. Gaillardon,
  ``Spinalflow: An architecture and dataflow tailored for spiking neural
  networks,'' in \emph{2020 ACM/IEEE 47th Annual International Symposium on
  Computer Architecture (ISCA)}.\hskip 1em plus 0.5em minus 0.4em\relax IEEE,
  2020, pp. 349--362.

\bibitem{lee2022parallel}
J.-J. Lee, W.~Zhang, and P.~Li, ``Parallel time batching: Systolic-array
  acceleration of sparse spiking neural computation,'' in \emph{2022 IEEE
  International Symposium on High-Performance Computer Architecture
  (HPCA)}.\hskip 1em plus 0.5em minus 0.4em\relax IEEE, 2022, pp. 317--330.

\bibitem{kim2012functional}
K.-H. Kim \emph{et~al.}, ``A functional hybrid memristor crossbar-array/cmos
  system for data storage and neuromorphic applications,'' \emph{Nano letters},
  2012.

\bibitem{ankit2017resparc}
A.~Ankit, A.~Sengupta, P.~Panda, and K.~Roy, ``Resparc: A reconfigurable and
  energy-efficient architecture with memristive crossbars for deep spiking
  neural networks,'' in \emph{Proceedings of the 54th Annual Design Automation
  Conference 2017}, 2017, pp. 1--6.

\bibitem{ni2017energy}
L.~Ni, Z.~Liu, W.~Song, J.~J. Yang, H.~Yu, K.~Wang, and Y.~Wang, ``An
  energy-efficient and high-throughput bitwise cnn on sneak-path-free digital
  reram crossbar,'' in \emph{2017 IEEE/ACM International Symposium on Low Power
  Electronics and Design (ISLPED)}.\hskip 1em plus 0.5em minus 0.4em\relax
  IEEE, 2017, pp. 1--6.

\bibitem{chakraborty2020pathways}
I.~Chakraborty, A.~Jaiswal, A.~Saha, S.~Gupta, and K.~Roy, ``Pathways to
  efficient neuromorphic computing with non-volatile memory technologies,''
  \emph{Applied Physics Reviews}, vol.~7, no.~2, p. 021308, 2020.

\bibitem{ankit2019puma}
A.~Ankit, I.~E. Hajj, S.~R. Chalamalasetti, G.~Ndu, M.~Foltin, R.~S. Williams,
  P.~Faraboschi, W.-m.~W. Hwu, J.~P. Strachan, K.~Roy \emph{et~al.}, ``Puma: A
  programmable ultra-efficient memristor-based accelerator for machine learning
  inference,'' in \emph{Proceedings of the Twenty-Fourth International
  Conference on Architectural Support for Programming Languages and Operating
  Systems}, 2019, pp. 715--731.

\bibitem{eyeriss}
Y.-H. Chen, T.~Krishna, J.~S. Emer, and V.~Sze, ``Eyeriss: An energy-efficient
  reconfigurable accelerator for deep convolutional neural networks,''
  \emph{IEEE journal of solid-state circuits}, vol.~52, no.~1, pp. 127--138,
  2016.

\bibitem{chen2018neurosim}
P.-Y. Chen, X.~Peng, and S.~Yu, ``Neurosim: A circuit-level macro model for
  benchmarking neuro-inspired architectures in online learning,'' \emph{IEEE
  Transactions on Computer-Aided Design of Integrated Circuits and Systems},
  vol.~37, no.~12, pp. 3067--3080, 2018.

\bibitem{plimpton2016crosssim}
S.~J. Plimpton, S.~Agarwal, R.~Schiek, and I.~Richter, ``Crosssim,'' Sandia
  National Lab.(SNL-NM), Albuquerque, NM (United States), Tech. Rep., 2016.

\bibitem{jain2020rxnn}
S.~Jain, A.~Sengupta, K.~Roy, and A.~Raghunathan, ``Rxnn: A framework for
  evaluating deep neural networks on resistive crossbars,'' \emph{IEEE
  Transactions on Computer-Aided Design of Integrated Circuits and Systems},
  vol.~40, no.~2, pp. 326--338, 2020.

\bibitem{krishnan2021siam}
G.~Krishnan, S.~K. Mandal, M.~Pannala, C.~Chakrabarti, J.-S. Seo, U.~Y. Ogras,
  and Y.~Cao, ``Siam: Chiplet-based scalable in-memory acceleration with mesh
  for deep neural networks,'' \emph{ACM Transactions on Embedded Computing
  Systems (TECS)}, vol.~20, no.~5s, pp. 1--24, 2021.

\bibitem{liang2021h2learn}
L.~Liang, Z.~Qu, Z.~Chen, F.~Tu, Y.~Wu, L.~Deng, G.~Li, P.~Li, and Y.~Xie,
  ``H2learn: High-efficiency learning accelerator for high-accuracy spiking
  neural networks,'' \emph{IEEE Transactions on Computer-Aided Design of
  Integrated Circuits and Systems}, 2021.

\bibitem{yin2022sata}
R.~Yin, A.~Moitra, A.~Bhattacharjee, Y.~Kim, and P.~Panda, ``Sata:
  Sparsity-aware training accelerator for spiking neural networks,''
  \emph{arXiv preprint arXiv:2204.05422}, 2022.

\bibitem{chakraborty2020geniex}
I.~Chakraborty, M.~F. Ali, D.~E. Kim, A.~Ankit, and K.~Roy, ``Geniex: A
  generalized approach to emulating non-ideality in memristive xbars using
  neural networks,'' in \emph{2020 57th ACM/IEEE Design Automation Conference
  (DAC)}.\hskip 1em plus 0.5em minus 0.4em\relax IEEE, 2020, pp. 1--6.

\bibitem{krishnan2021robust}
G.~Krishnan, J.~Sun, J.~Hazra, X.~Du, M.~Liehr, Z.~Li, K.~Beckmann, R.~V.
  Joshi, N.~C. Cady, and Y.~Cao, ``Robust rram-based in-memory computing in
  light of model stability,'' in \emph{2021 IEEE International Reliability
  Physics Symposium (IRPS)}.\hskip 1em plus 0.5em minus 0.4em\relax IEEE, 2021,
  pp. 1--5.

\bibitem{krishnan2022exploring}
G.~Krishnan, L.~Yang, J.~Sun, J.~Hazra, X.~Du, M.~Liehr, Z.~Li, K.~Beckmann,
  R.~Joshi, N.~C. Cady \emph{et~al.}, ``Exploring model stability of deep
  neural networks for reliable rram-based in-memory acceleration,'' \emph{IEEE
  Transactions on Computers}, 2022.

\bibitem{truong2014new}
S.~N. Truong and K.-S. Min, ``New memristor-based crossbar array architecture
  with 50-\% area reduction and 48-\% power saving for matrix-vector
  multiplication of analog neuromorphic computing,'' \emph{JSTS: Journal of
  Semiconductor Technology and Science}, vol.~14, no.~3, pp. 356--363, 2014.

\bibitem{bhattacharjee2022examining}
A.~Bhattacharjee, Y.~Kim, A.~Moitra, and P.~Panda, ``Examining the robustness
  of spiking neural networks on non-ideal memristive crossbars,'' \emph{arXiv
  preprint arXiv:2206.09599}, 2022.

\bibitem{cifar}
A.~Krizhevsky, G.~Hinton \emph{et~al.}, ``Learning multiple layers of features
  from tiny images,'' 2009.

\bibitem{le2015tiny}
Y.~Le and X.~Yang, ``Tiny imagenet visual recognition challenge,'' \emph{CS
  231N}, vol.~7, no.~7, p.~3, 2015.

\bibitem{shafiee2016isaac}
A.~Shafiee, A.~Nag, N.~Muralimanohar, R.~Balasubramonian, J.~P. Strachan,
  M.~Hu, R.~S. Williams, and V.~Srikumar, ``Isaac: A convolutional neural
  network accelerator with in-situ analog arithmetic in crossbars,'' \emph{ACM
  SIGARCH Computer Architecture News}, vol.~44, no.~3, pp. 14--26, 2016.

\bibitem{diehl2015unsupervised}
P.~U. Diehl and M.~Cook, ``Unsupervised learning of digit recognition using
  spike-timing-dependent plasticity,'' \emph{Frontiers in computational
  neuroscience}, 2015.

\bibitem{sengupta2019going}
A.~Sengupta, Y.~Ye, R.~Wang, C.~Liu, and K.~Roy, ``Going deeper in spiking
  neural networks: Vgg and residual architectures,'' \emph{Frontiers in
  neuroscience}, vol.~13, p.~95, 2019.

\bibitem{diehl2015fast}
P.~U. Diehl, D.~Neil, J.~Binas, M.~Cook, S.-C. Liu, and M.~Pfeiffer,
  ``Fast-classifying, high-accuracy spiking deep networks through weight and
  threshold balancing,'' in \emph{2015 International joint conference on neural
  networks (IJCNN)}.\hskip 1em plus 0.5em minus 0.4em\relax ieee, 2015, pp.
  1--8.

\bibitem{han2020deep}
B.~Han and K.~Roy, ``Deep spiking neural network: Energy efficiency through
  time based coding,'' in \emph{Computer Vision--ECCV 2020: 16th European
  Conference, Glasgow, UK, August 23--28, 2020, Proceedings, Part X 16}.\hskip
  1em plus 0.5em minus 0.4em\relax Springer, 2020, pp. 388--404.

\bibitem{li2021free}
Y.~Li, S.~Deng, X.~Dong, R.~Gong, and S.~Gu, ``A free lunch from ann: Towards
  efficient, accurate spiking neural networks calibration,'' \emph{arXiv
  preprint arXiv:2106.06984}, 2021.

\bibitem{rueckauer2017conversion}
B.~Rueckauer, I.-A. Lungu, Y.~Hu, M.~Pfeiffer, and S.-C. Liu, ``Conversion of
  continuous-valued deep networks to efficient event-driven networks for image
  classification,'' \emph{Frontiers in neuroscience}, vol.~11, p. 682, 2017.

\bibitem{wu2018spatio}
Y.~Wu, L.~Deng, G.~Li, J.~Zhu, and L.~Shi, ``Spatio-temporal backpropagation
  for training high-performance spiking neural networks,'' \emph{Frontiers in
  neuroscience}, vol.~12, p. 331, 2018.

\bibitem{kim2022rate}
Y.~Kim, H.~Park, A.~Moitra, A.~Bhattacharjee, Y.~Venkatesha, and P.~Panda,
  ``Rate coding or direct coding: Which one is better for accurate, robust, and
  energy-efficient spiking neural networks?'' in \emph{ICASSP 2022-2022 IEEE
  International Conference on Acoustics, Speech and Signal Processing
  (ICASSP)}.\hskip 1em plus 0.5em minus 0.4em\relax IEEE, 2022, pp. 71--75.

\bibitem{bntt}
Y.~Kim and P.~Panda, ``Revisiting batch normalization for training low-latency
  deep spiking neural networks from scratch,'' \emph{Frontiers in
  Neuroscience}, 2021.

\bibitem{liu2015vortex}
B.~Liu, H.~Li, Y.~Chen, X.~Li, Q.~Wu, and T.~Huang, ``Vortex: Variation-aware
  training for memristor x-bar,'' in \emph{Proceedings of the 52nd Annual
  Design Automation Conference}, 2015, pp. 1--6.

\bibitem{bhattacharjee2021neat}
A.~Bhattacharjee, L.~Bhatnagar, Y.~Kim, and P.~Panda, ``Neat: Non-linearity
  aware training for accurate, energy-efficient and robust implementation of
  neural networks on 1t-1r crossbars,'' \emph{IEEE Transactions on
  Computer-Aided Design of Integrated Circuits and Systems}, 2021.

\bibitem{bhattacharjee2021efficiency}
A.~Bhattacharjee, A.~Moitra, and P.~Panda, ``Efficiency-driven hardware
  optimization for adversarially robust neural networks,'' in \emph{2021
  Design, Automation \& Test in Europe Conference \& Exhibition (DATE)}.\hskip
  1em plus 0.5em minus 0.4em\relax IEEE, 2021, pp. 884--889.

\bibitem{bhattacharjee2021switchx}
A.~Bhattacharjee and P.~Panda, ``Switchx: Gmin-gmax switching for
  energy-efficient and robust implementation of binary neural networks on reram
  xbars,'' 2021.

\bibitem{royfundamental}
S.~K. Roy, A.~Patil, and N.~R. Shanbhag, ``Fundamental limits on the
  computational accuracy of resistive crossbar-based in-memory architectures.''

\bibitem{jaiswal20198t}
A.~Jaiswal, I.~Chakraborty, A.~Agrawal, and K.~Roy, ``8t sram cell as a
  multibit dot-product engine for beyond von neumann computing,'' \emph{IEEE
  Transactions on Very Large Scale Integration (VLSI) Systems}, vol.~27,
  no.~11, pp. 2556--2567, 2019.

\bibitem{hajri2019rram}
B.~Hajri, H.~Aziza, M.~M. Mansour, and A.~Chehab, ``Rram device models: A
  comparative analysis with experimental validation,'' \emph{IEEE Access},
  vol.~7, pp. 168\,963--168\,980, 2019.

\bibitem{jiang2013detailed}
N.~Jiang, D.~U. Becker, G.~Michelogiannakis, J.~Balfour, B.~Towles, D.~E. Shaw,
  J.~Kim, and W.~J. Dally, ``A detailed and flexible cycle-accurate
  network-on-chip simulator,'' in \emph{2013 IEEE international symposium on
  performance analysis of systems and software (ISPASS)}.\hskip 1em plus 0.5em
  minus 0.4em\relax IEEE, 2013, pp. 86--96.

\end{thebibliography}


\end{document}